\pdfoutput=1
\documentclass[journal,onecolumn]{IEEEtran}
\usepackage[utf8]{inputenc} 
\usepackage{color}
\usepackage[T1]{fontenc}    
\usepackage{hyperref}       
\usepackage{url}            
\usepackage{makecell}
\usepackage{amsmath,amsfonts,amsthm,bm} 
\usepackage{booktabs}       
\usepackage{amsfonts}       
\usepackage{nicefrac}       
\usepackage{lipsum}
\usepackage{ragged2e}
\usepackage{setspace}
\usepackage{array}
\usepackage{multirow}
\usepackage{microtype}
\usepackage{graphicx}
\usepackage[english]{babel}
\usepackage{thmtools}
\usepackage{thm-restate}
\usepackage[font=small,labelfont=bf]{caption}
\def\etal{\textit{et al. }}
\def\sep{, }
\newcommand{\han}[1]{{\color{black}{#1}}}
\newcommand{\B}[1]{{\textbf{#1}}}
\newtheorem{theorem}{Theorem}
\newenvironment{keyword}{\begin{IEEEkeywords}}{\end{IEEEkeywords}}

\begin{document}
\title{A Learning-based Framework for Topology-Preserving Segmentation using Quasiconformal Mappings}
\author{Han Zhang, Lok Ming Lui
\thanks{H. Zhang is with the City University of Hong Kong, Hong Kong SAR(e-mail:hzhang863-c@my.cityu.edu.hk).}
\thanks{L.M. Lui is with the Chinese University of Hong Kong, Hong Kong SAR(e-mail: lmlui@math.cuhk.edu.hk). Corresponding author.}}
\maketitle
\begin{abstract}
We propose the Topology-Preserving Segmentation Network, a deformation-based model that can extract objects in an image while maintaining their topological properties. This network generates segmentation masks that have the same topology as the template mask, even when trained with limited data. The network consists of two components: the Deformation Estimation Network, which produces a deformation map that warps the template mask to enclose the region of interest, and the Beltrami Adjustment Module, which ensures the bijectivity of the deformation map by truncating the associated Beltrami coefficient based on Quasiconformal theories. The proposed network can also be trained in an unsupervised manner, eliminating the need for labeled training data. This is achieved by incorporating an unsupervised segmentation loss. Our experimental results on various image datasets show that TPSN achieves better segmentation accuracy than state-of-the-art models with correct topology. Furthermore, we demonstrate TPSN's ability to handle multiple object segmentation.
\end{abstract}
\begin{keyword}
Topology-Preserving\sep Prior Knowledge\sep Deformable Model\sep Medical Image Segmentation
\end{keyword}
\section{Introduction}
Image segmentation, the process of identifying and separating regions of interest in images, is a crucial task in many applications such as object recognition, image editing and medical imaging. For instance, medical image segmentation is known to be a crucial step in many clinical applications, including disease diagnosis and monitoring of disease progression. While deep learning has revolutionized the field of image processing in recent years, developing learning-based segmentation models is still facing a significant challenge due to the limited availability of the training data. This situation is particularly common in the medical field, as medical image data is frequently subject to privacy and legal regulations that limit its availability. The lack of data poses challenges to training models that are capable of achieving high levels of accuracy in segmenting anatomical structures from medical images.

However, despite the limited availability of image data, it is worth noting that structures of interest in many applications follow strict morphological rules. These rules can be incorporated as prior knowledge into learning-based segmentation models. By using these prior rules, it is possible to guide the segmentation process and achieve better segmentation quality, even with limited data availability. For instance, anatomical structures such as the kidney or liver are known to be simply-connected, while vessel walls are typically circularly shaped. Integrating the topological rule into a learning-based segmentation model can improve the segmentation result, as well as ensure that the segmented object is consistent with the anatomical structure being imaged. 

Conventional segmentation methods often fail to consider geometric information, which can result in inaccurate segmentation with topological errors. For example, mathematical models for image segmentation based on pixel-wise classification may produce isolated segmented regions and create holes in anatomical structures. These issues can be clinically problematic and impractical \cite{chan2005level}. To address this problem, recent works have incorporated prior shape information into regularization terms, which can improve segmentation results but do not ensure topological correctness \cite{chan2005level}. On the other hand, deformation-based segmentation models can enforce topological and geometric constraints by warping a template image to a target domain that encompasses the region of interest. For instance, topology- and convexity-preserving segmentation models have been proposed recently. These models deform a template shape by mapping it to segment the image \cite{chan2018topology,siu2020image}. The properties of the mapping are carefully controlled. 

In recent years, learning-based imaging models have gained significant attention as a means to integrate data information and enhance efficiency. By utilizing U-Net\cite{ronneberger2015u}, dense spatial transformations can be generated using deep neural network models\cite{jaderberg2015spatial}. This opens up possibilities for developing learning-based segmentation techniques based on deformable models. However, enforcing geometric constraints on the deformation map within a deep neural network remains a challenge. Recent studies have explored various regularizations to be imposed on the output of a spatial transformer network or its sub-layers. However, these approaches often tend to be over-constrained, leading to sub-optimal results \cite{wyburd2021teds}, or lack mathematical justifications to guarantee a solution that satisfies the prescribed geometric conditions \cite{lee2019tetris}.

In this paper, we present a topology-preserving segmentation framework that utilizes a learning-based approach. Our proposed framework, namely Topology-Preserving Segmentation Network (TPSN), learns to segment an input image by deforming the template mask to enclose the region of interest while preserving topology. The preservation of topology is achieved by ensuring the bijectivity of the deformation map. To enforce the bijectivity of the deformation map, we introduce the Deformation Estimation Network regularized by ReLU-Jacobian and the Beltrami Adjustment Module based on Quasiconformal theories. The ReLU-Jacobian regularizer, which is the negative of the Jacobian determinant activated by the ReLU function, discourages changes in orientation under the deformation map in the segmentation model, thereby promoting bijectivity. To guarantee bijectivity, we further propose a Beltrami adjustment module based on quasiconformal theories that adjust the Beltrami coefficient of the mapping, which measures local geometric distortions. This module can guarantee the bijectivity of the deformation map, even in extreme cases, resulting in topology-preserving segmentation results. Additionally, our proposed network can be trained in an unsupervised manner, which eliminates the need for labeled training data. This is achieved by incorporating an unsupervised loss into our proposed model.

We conduct extensive experiments to evaluate the performance of our proposed framework. The proposed model is first applied for supervised segmentation, which involves training the segmentation network with a set of images with labeled masks. The loss function includes the proposed ReLU Jacobian regularization and a pixel-wise segmentation loss, such as the Dice loss. Our proposed network successfully produces segmentation results with the prescribed topological constraint. We evaluate the performance of the model on the KiTS21\cite{heller2021state} and ACDC\cite{bernard2018deep} dataset for simply-connected and doubly-connected object segmentation, respectively. To segment more general multiply-connected objects, we design the Fill First, Dig Second(FFDS) strategy. To further improve the segmentation accuracy, we propose a multi-level TPSN architecture that predicts the segmentation mask from a coarse-to-fine fashion. We also test our proposed model for unsupervised segmentation using an unsupervised segmentation loss in the model. Our proposed framework is evaluated on the BTCV\cite{btcv2015} dataset extracted from MICCAI 2015 challenge, and experimental results are also promising. From the experiments on the corrupted images, our model is robust under extreme cases where very noisy images are considered, with the help of the Beltrami adjustment module.

To summarize, the contributions of this work are as follows:
\begin{itemize}
    \item We proposed a segmentation framework called Topology-Preserving Segmentation Network (TPSN), which learns to segment an input image by deforming the template mask to enclose the region of interest while preserving topology.
    \item We propose a methodology for topology-preserving segmentation that combines the ReLU-Jacobian regularizer and the Beltrami adjustment module. The ReLU-Jacobian regularizer promotes bijectivity and the Beltrami adjustment module guarantees bijectivity even in extreme and noisy cases. This approach results in topology-preserving segmentation masks
    \item Our proposed network can be trained in an unsupervised manner using the Mumford-Shah model's fidelity term, which distinguishes the foreground and background based on image intensities, eliminating the need for labeled training data. 
\end{itemize}

\section{Related Work}

\subsection{Image Segmentation}
Image segmentation has been studied extensively and various segmentation models have been developed in recent years. The active contour model \cite{kass1988snakes} is perhaps the earliest segmentation model, which evolves a parameterized curve iteratively to the object boundary. Later, Chan \etal \cite{chan2001active} proposed the active contour model without edge, based on the Mumford-Shah functional and level sets, to segment an image by curve evolution. The model has an advantage of allowing topological changes. With the popularity of deep learning in recent years, a lot of segmentation models using deep architectures have been invented. Ronneberger \etal \cite{ronneberger2015u} proposed the learning-based segmentation model using the U-Net structure and achieved impressive results. The extension of U-Net for 3D volumetric images was later developed in \cite{milletari2016v,cciccek20163d}. As learning-based segmentation models require data, multiple works have been devoted to generating meaningful training data. A series of dataset \cite{tschandl2018ham10000,bernard2018deep,heller2021state,btcv2015} are organized and open to the public for research purposes. Some works design learning-based segmentation models by manipulating data. For instance, a segmentation framework, called nnUNet, was proposed in \cite{isensee2021nnu}, which uses only a standard U-Net architecture with a series of practical pre- and post- processing strategies on the images. This framework achieved outstanding results on multiple segmentation datasets. More learning-based models have been invented recently by adjusting the network architectures. Jha \etal \cite{jha2020doubleu} proposed to use one UNet to predict an initial mask and apply an additional one to refine the segmentation result. 

Besides, unsupervised learning-based segmentation models have also been considered, which train the model by images without labeled masks as the ground truth. Ouali \etal\cite{ouali2020autoregressive} proposed the autoregressive unsupervised image segmentation model by maximizing the mutual information between different constructed views.

\subsection{Deformable Model}
Deformable models for image segmentation have also been explored by different research groups. The main idea of deformable segmentation models is to obtain the result by searching for a suitable deformation. The active contour model \cite{kass1988snakes} can be considered as a deformable model, which deforms a collection of points discretizing a curve to enclose the boundary of the object. Cootes \etal \cite{cootes1995active} extended the active contour model to a learnable model by extracting the main pattern of variations using principle component analysis (PCA). More recent deformable image segmentation models consider a dense spatial deformation map between the template image and the target image. The template mask is then deformed to the shape of the object in the image. Chen \etal \cite{chen2021generalized} proposed a new dual-front scheme, based on asymmetric quadratic metrics, which integrates image features and a vector field derived from the evolving contour. Chan \etal \cite{chan2018topology} proposed a deformation-based segmentation model using quasiconformal maps and the segmentation results are guaranteed to be topology-preserving. Siu \etal \cite{siu2020image} applied the dihedral angle in the deformable model for image segmentation to incorporate the partial convexity and topology constraints. Zhang \etal \cite{zhang2021topology} proposed a deformable model using the hyperelastic regularization, which gives rise to a topology-preserving segmentation model and can be applied to 3D volumetric images. Convexity prior is further integrated into the model in \cite{zhang2021topoconv}. Learning-based models that utilize deformable models have gained significant attention in recent years \cite{dinsdale2019spatial, zeng2019liver} since the introduction of spatial transformer networks \cite{jaderberg2015spatial}. To improve the preservation of topology in the final output, Lee \etal \cite{lee2019tetris} applied Laplacian regularization. Wyburd \etal \cite{wyburd2021teds} proposed a topology-preserving deep segmentation network by compositing a series of bijective deformations. However, these methods may not have a mathematical guarantee of topological preservation or may produce suboptimal results due to over-constraints, particularly for structures with complex geometry.

\subsection{Prior Information In Neural Network}
Recent research has focused on incorporating prior information into deep neural networks by designing an appropriate loss function that can integrate prior knowledge to regulate predicted masks. Yan \etal \cite{yan2020convexity} proposed a convexity-preserving method for variational level-set-based image segmentation. Mosinska \etal \cite{mosinska2018beyond} defined a topological loss that considers higher-order topological features of linear structures. Zhou \etal \cite{zhou2019prior} proposed a segmentation model that statistically considers the relative location and size, performing well in multi-organ segmentation. Hu \etal \cite{hu2019topology} designed a continuous-valued loss function to enforce topological correctness using computational topology theories. Shit \etal \cite{shit2021cldice} introduced clDice to measure the Dice of skeletons between the prediction and the ground truth for tubular structures. Clough \etal \cite{clough2020topological} proposed using differentiable persistent homology to drive the segmentation process to produce an output with a prescribed Betti number.
\section{Proposed Method}

In this section, we describe our proposed topology-preserving segmentation network (TPSN) in detail. 

\subsection{Overall framework}

\begin{figure*}
    \centering
    \includegraphics[width = \textwidth]{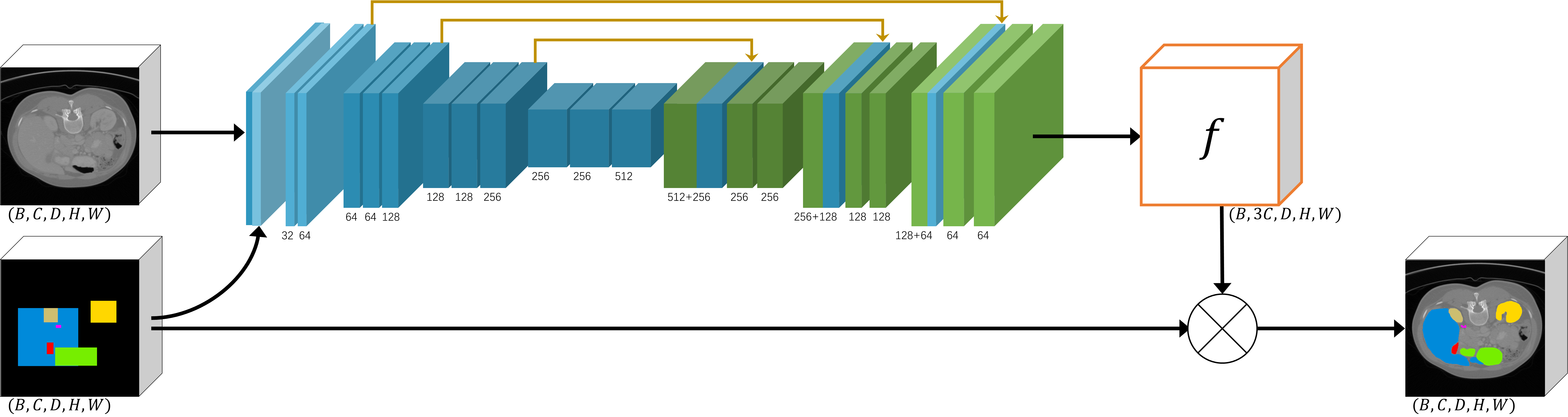}
    \caption{The TPSN network architecture consists of two components: the Deformation Estimation Network (represented by the yellow box) and the Beltrami Adjustment Module (represented by the blue box). For a segmentation task for $q$ classes of objects, the Deformation Estimation Network utilizes an encoder-decoder architecture that takes in a prior $M_{temp}$ and an image $I$ with $C$ channels, and produces $q$ mappings for each class. This mapping is then used to deform prior masks to enclose regions of interest $M_{pred}$. The component is designed with ReLU-Jacobian regularization, which promotes the output of a topology-preserving mapping. The Beltrami Adjustment Module consists of two convolutional layers activated by a $Tanh$. This component ensures that the mapping is topology-preserving by truncating $\mu$ into $\Tilde{\mu}$. The Beltrami Solver Network \cite{chen2021deep} reconstructs the quasiconformal mapping $\Tilde{f}$ that corresponds to the truncated $\Tilde{\mu}$. The corresponding prediction mask $\widetilde{M}_{pred}$ is then obtained.}
    \label{fig:framework}
\end{figure*}

Our proposed TPSN is a deep neural network that takes an image and a template mask as input and outputs the segmentation mask. Figure \ref{fig:framework} depicts the overall framework, which is composed of two components: (1) the Deformation Estimation Network (DEN) and (2) the Beltrami Adjustment Module (BAM). 

The DEN takes the image and the template mask as input, where the template mask encodes the desired topological properties. Based on the input, a deformation mapping used to warp the template mask into the region of interest is generated. The ReLU-Jacobian term is included in the loss function of the DEN as a soft regularisation term to encourage the deformation mapping to be bijective. Under normal circumstances, training the DEN with the ReLU-Jacobian regularization enables it to generate bijective mappings. However, in challenging situations like working with noisy images, there is no assurance that the deformation mapping produced by the DEN will maintain bijectivity. To tackle this challenge, the Beltrami Adjustment Module is introduced as a solution to ensure the bijectivity of the generated mapping. This module, which functions as a post-processing operation, does not learn from data. It is optimized by minimizing \eqref{eq:truncation}, which relies solely on the predicted mapping and the corresponding transformed mask obtained from DEN. By employing this approach, BAM is able to produce a bijective quasi-conformal mapping for each input while preserving the region enclosed by DEN to the greatest extent possible.

Finally, the bijective quasiconformal map is used to warp the template mask to construct the final segmentation result, which is guaranteed topology-preserving. In the following subsection, we will explain every component in our framework.

\subsection{Deformation Estimation Network (DEN)}
\subsubsection*{Network architecture} The architecture of the Deformation Estimation Network (DEN) is depicted in Figure \ref{fig:framework}, where the channels are delineated beneath each layer. Each convolutional operation within the network is performed using a $3\times 3$ kernel for 2D images and a $3\times 3\times 3$ kernel for 3D images. Subsequently, batch normalization is applied, followed by activation through the rectified linear unit (\textit{ReLU}) function. Throughout the process, three max-pooling and three bilinear upsampling operations are incorporated for feature extraction and resolution enhancement, respectively. The network takes the image $I$ to be segmented and concatenates it with a pre-defined template mask $M_{temp}$, which is chosen based on the topological prior of the object being segmented. This concatenated input is then fed through an encoder-decoder network, which generates a deformation mapping $f=\mathcal{N}(I,M_{temp})$ that warps the template mask into a predicted mask $M_{pred}=M_{temp}\circ f$. To ensure that the orientation of the deformation mapping is preserved as much as possible, a suitable loss function is minimized during the training process. The specific details of this loss function will be described in the next paragraph.

\subsubsection*{Loss} The main goal of our segmentation framework is to obtain an accurate segmentation result that preserves the prescribed topology. To achieve this, our strategy is to obtain a bijective deformation mapping that transforms a template mask with the prescribed topological prior to obtain an accurate segmentation mask. The bijectivity of the deformation mask ensures that the topology of the segmentation mask is consistent with that of the template mask. To accomplish this, a suitable loss function is necessary. In this work, we propose a loss function that consists of the fidelity term and the regularization term. The regularization term aims to regularize the deformation mapping. To encourage the bijectivity of the deformation mapping, the ReLU Jacobian regularizer is proposed. A map is homeomorphic if its Jacobian is positive everywhere. Motivated by this, the following ReLU Jacobian regularizer is proposed:
\begin{equation}
\mathcal{L}_{Jac}(f) = || ReLU ( - \det\nabla(f)) ||_1
\label{eq:relujac}
\end{equation}
where $\det\nabla(f)$ denotes the Jacobian determinant of $f$, $ReLU(\cdot)$ represents the ReLU activation function that sets negative values to zero while keeping positive values. $\mathcal{L}_{Jac}(f)$ penalizes the portion of the deformation map with negative Jacobian. Minimizing this loss function encourages the deformation mapping to preserve the orientation as much as possible.

Besides, we introduce an additional Laplacian regularization term aimed at smoothing the deformation mapping:
\begin{equation}
\mathcal{L}_{Lap}(f) = || \Delta(f) ||_1
\end{equation}
where $\Delta(f)$ denotes the Laplacian of the mapping $f$. This Laplacian regularization promotes smoothness in the deformation field, thereby mitigating abrupt changes and preserving the overall structural integrity of the segmented objects. By incorporating this regularization technique alongside the ReLU-Jacobian regularization, we enhance the accuracy of the segmentation output.

Next, the fidelity loss aims to promote segmentation results that accurately match and faithfully represent the input image data. For supervised learning where the training data has labeled masks, the Dice loss is used. The overall loss function is then given by:
\begin{equation}
\mathcal{L} = \mathcal{L}_{Dice}(M_{pred},M_{label})
    +\lambda_{Jac} \mathcal{L}_{Jac}(f)
    +\lambda_{Lap} \mathcal{L}_{Lap}(f)
    \label{eq:suploss}
\end{equation}
where $\mathcal{L}_{Dice}$ is the Dice loss between the predicted mask $M_{pred}$ and label mask $M_{label}$. $\lambda_{Jac}$ and $\lambda_{Lap}$ are the weighting parameters for $\mathcal{L}_{Jac}(f)$ and $\mathcal{L}_{Lap}(f)$ respectively. \han{It's also worth to note that the label mask $M_{label}$ are given while the predicted mask $M_{pred}$ are transformed from a given template mask $M_{temp}$ by the predicted mapping $f$.}

For unsupervised learning without the need for labeled training data, the following unsupervised segmentation loss is used as the fidelity term:
\begin{equation} 
\mathcal{L}_{seg}(I,M_{pred}) = \int_{D}(I - c_1 M_{pred} - c_2 (1-M_{pred}))^2 d\boldsymbol{x}
\label{eq:CV}
\end{equation}
where $c_1=\frac{\int_{D} I \cdot M_{pred}d\boldsymbol{x}}{\int_{D} M_{pred}d\boldsymbol{x}}$ and $c_2=\frac{\int_{D} I \cdot (1-M_{pred})d\boldsymbol{x}}{\int_{D} 1-M_{pred}d\boldsymbol{x}}$, $D$ is the whole image domain. In fact, $c_1$ denotes the mean pixel value of the foreground (region covered by the predicted mask) while $c_2$ is the mean pixel value of the background. 
$M_{pred} = M_{temp} \circ f$ represents the final segmentation mask. $M_{pred}$ is derived by deforming the binary template mask $M_{temp}$ using the mapping $f$. Consequently, with nearest sampling method, $M_{pred}$ is also a binary image, where the interior region has an intensity value of 1, and the exterior region has an intensity value of 0. The purpose of Equation \eqref{eq:CV} is to generate a segmentation mask that ensures the intensity values of the input image $I$ within the interior region of the mask are close to the average intensity of that region. Similarly, the intensity values of $I$ within the exterior region of the mask should be close to the average intensity of the exterior region. This objective aims to enable accurate segmentation of a piecewise constant image, where the segmentation mask aligns well with regions of consistent intensity. Thus, the overall loss function for the unsupervised case is given by:
\begin{equation} 
\mathcal{L} = \mathcal{L}_{seg}(I,M_{pred})
    +\lambda_{Jac} \mathcal{L}_{Jac}(f)
    +\lambda_{Lap} \mathcal{L}_{Lap}(f)
\end{equation}

\han{Overall, given a template mask $M_{temp}$ and label mask $M_{true}$. The predicted mask can be computed from $M_{pred}=M_{temp}\circ f$, where $f$ is the predicted output from a deformation estimation network.}

\subsection{Beltrami Adjustment Module (BAM)}
\begin{figure}
    \centering
    \includegraphics[width=0.5\textwidth]{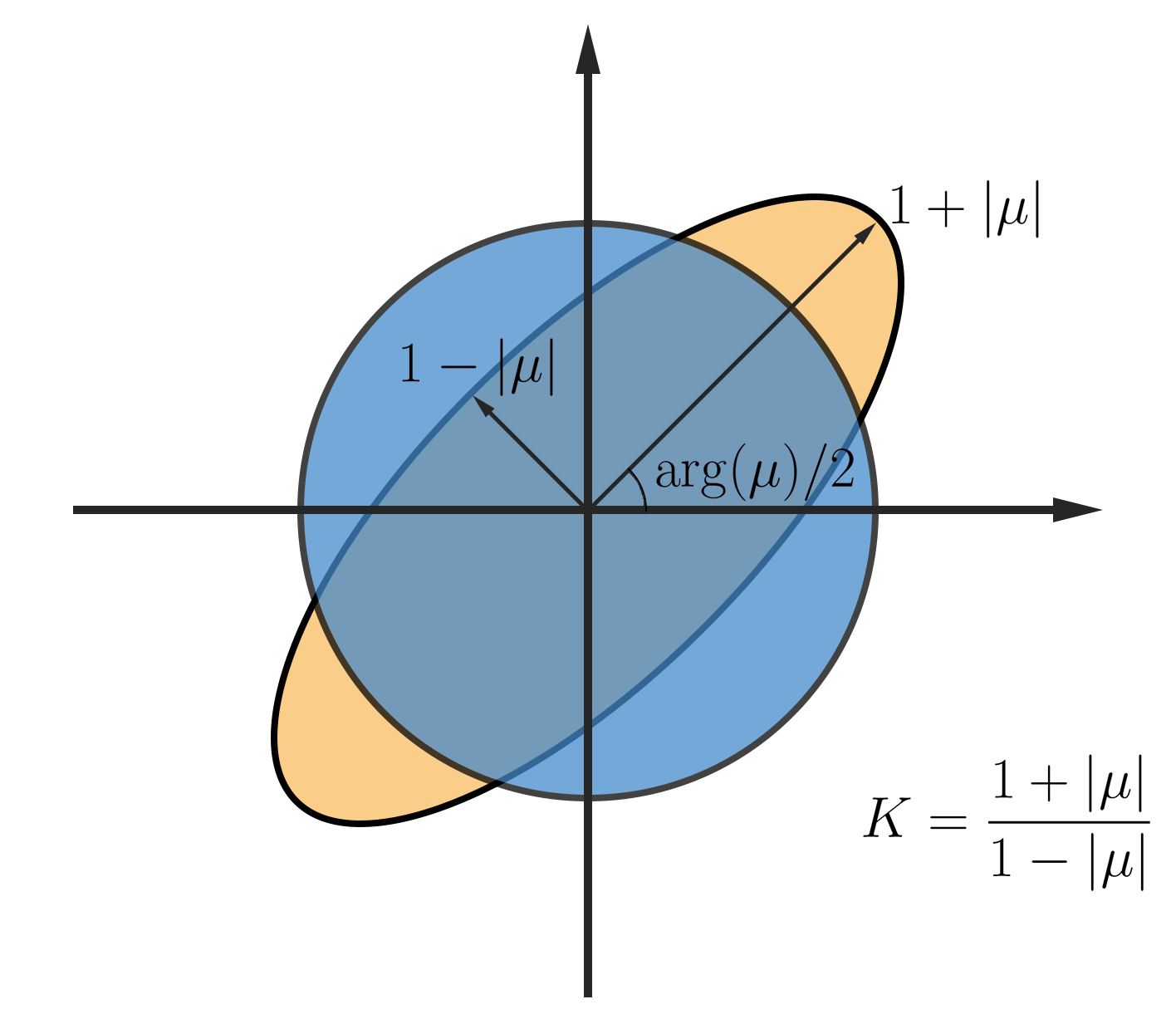}
    \caption{Quasi-conformal maps infinitesimal circles to ellipses. The Beltrami coefficient measures the distortion or dilation of the ellipse under the QC map.}
    \label{fig:qcmap}
\end{figure}

With the ReLU Jacobian regularizer, DEN can produce a deformation map that preserves the orientation as much as possible. In the case where the input data is close to the distribution of the training dataset, the output deformation mapping from DEN is mostly bijective, resulting in a topology-preserving segmentation mask. However, when a new input is highly noisy, which makes it significantly different from the training data distribution, the resulting mapping may not be perfectly bijective. To address this issue and ensure that the deformation mapping produced by our segmentation framework is bijective, we propose the implementation of a Beltrami Adjustment Module (BAM). This module adjusts the deformation mapping to ensure bijectivity. Every mapping is associated with a geometric quantity called the Beltrami coefficient, which is defined as follows: 
\begin{equation}\label{mu}
\mu=\frac{\partial f}{\partial \bar{z}} / \frac{\partial f}{\partial z},
\end{equation}
\noindent where $\frac{\partial}{\partial z} = \frac{1}{2}(\frac{\partial}{\partial x} -{i}\frac{\partial}{\partial y})$ and $\frac{\partial}{\partial \bar{z}} = \frac{1}{2}(\frac{\partial}{\partial x} +{i}\frac{\partial}{\partial y})$. ${i}$ is the imaginary unit.

Based on $\mu$, it is possible to determine that the angle of maximum magnification is $arg(\mu)/2$, accompanied by a magnification factor of $|\frac{\partial f}{\partial z}| (1+|\mu|)$. Similarly, the angle of maximum contraction is the orthogonal angle $(arg(\mu) - \pi)/2$, with a contraction factor of $|\frac{\partial f}{\partial z}| (1-|\mu|)$ \cite{lam2014landmark}. This implies that the Beltrami coefficient $\mu$ measures the local geometric distortion under $f$, as illustrated in Figure \ref{fig:qcmap}.

The Beltrami coefficient $\mu$ is a complex-valued function defined on the image domain. A Beltrami coefficient $\mu$ in 2 dimensions can be expressed as a 2-channel image of the same size as the corresponding input images. In the discrete case, the image domain is triangulated and each deformation mapping is treated as a piecewise linear function over each triangular face. The first derivatives of a piecewise linear function are constant on each triangular face. Consequently, the Beltrami coefficient can be regarded as a complex-valued function defined over each triangular face. The discretization details will be described in the next section. $\mu$ measures the local geometric distortion under the mapping. If $\mu(T) = 0$ on a triangular face $T$, the three angles of the triangle will be preserved under the deformation. 

Importantly, the magnitude of the Beltrami coefficient is strictly less than 1 everywhere for an orientation-preserving homeomorphism. Leveraging this fact, the Beltrami Adjustment Module (BAM) is designed to control geometric distortion by adjusting the Beltrami coefficient associated with the deformation mapping. After obtaining $f$ from the DEN, we calculate its corresponding Beltrami coefficient, $\mu$, using equation \ref{mu}. If a region has undergone a flip resulting in an orientation change, $|\mu|>1$. Our goal is to modify $\mu$ to $\Tilde{\mu}$ such that the associated mapping with $\Tilde{\mu}$ is similar to $f$, and $|\tilde{\mu}|<1$ throughout the image domain. To achieve this, we input $\mu$ into a truncation module $F$, which returns a truncated $\Tilde{\mu} = F(\mu)$ with a magnitude less than one everywhere in the image domain.

\han{
To enforce the Dirichlet boundary condition for the deformation $\Tilde{f}$ output by the BAM in the image domain $D$, the BSNet is designed to output a mapping that satisfies $\Tilde{f}(x)=x$ on the boundary of $D$. Specifically, the BAM first outputs a Beltrami coefficient $\Tilde{\mu}$ with a supremum norm strictly less than $1$ to ensure bijectivity. Given this Beltrami coefficient, the BSNet outputs a quasiconformal mapping whose boundary map is an identity map. This is achieved by setting the BSNet to freeze the position of points on the boundary. 
}

The truncation module $F$ is a simple network comprising two convolution layers and a Tanh activation function that restricts the output's norm to be less than $1$. However, this truncation might cause the associated quasiconformal mapping to deviate from the original map and degrade the final mask. Therefore, we initialize $F$ by optimizing its parameters according to the loss function:
\begin{equation}
    \mathcal{L}_{init} = ||\mu - \Tilde{\mu}||,
    \label{eq:initialize}
\end{equation}
and introduce an additional loss term to regulate the truncation module $F$. This loss term is given by:
\begin{equation}
    \mathcal{L}_{trun} = ||M_{temp} \circ f - M_{temp} \circ \Tilde{f}|| + \lambda_{\mu} ReLU(|\Tilde{\mu}|^2 - 1),
    \label{eq:truncation}
\end{equation}
where $\Tilde{f}$ is the mapping generated by the truncated coefficient and $M_{temp}$ is a template mask used for training. The first term prevents the final mapping from deviating too far from the predicted mapping of our DEN. The second term, as well as the activation function, ensures that the truncated $\Tilde{\mu}$ is less than $1$. The condition $\Tilde{\mu}(T)<1$ for all triangular faces $T$ ensures the bijectivity of the mapping $\Tilde{f}$, which is guaranteed by the following theorem: 

\medskip

\begin{theorem}
\label{thm}
The Beltrami Adjustment Module outputs a bijective mapping $\Tilde{f}$ if and only if its associated Beltrami coefficient $\Tilde{\mu}$ is well-defined and satisfies $|\Tilde{\mu}(T)|<1$ for all triangular faces $T$.
\end{theorem}
\begin{proof}
Denote the image domain as $\Omega$. $\Tilde{f}$ is piecewise linear on every triangular face $T$. In $T$, $\Tilde{f}$ can be written as $a^x_1 x + a^x_2 y + b^x + i(a^y_1 x + a^y_2 y + b^y)$.
The Jacobian $J_T$ of $\Tilde{f}$ is a constant on each $T$. Also, $J_T = (a_1^x a_2^y - a_1^y a_2^x) = \frac{1}{4}\left( (a_1^x + a_2^y)^2 + (a_2^x - a_1^y)^2 \right) - \frac{1}{4}\left( (a_1^x - a_2^y)^2 + (a_2^x + a_1^y)^2 \right)   =  |\frac{\partial \Tilde{f}}{\partial z}(T)|^2 (1-|\Tilde{\mu}(T)|^2)$. If $\Tilde{f}$ is bijective and orientation-preserving, the Jacobian of $\Tilde{f}$ is positive everywhere and $|\frac{\partial \Tilde{f}}{\partial z}(T)|>0 $ for all $T$ since $\Tilde{\mu}$ is well-defined. Also, $(1-|\Tilde{\mu}(T)|^2)>0$ and $|\Tilde{\mu}(T)|<1$ for all $T$. 
Conversely, since $\Tilde{\mu}(T) = \frac{\partial \Tilde{f}}{\partial\Bar{z}}(T) / \frac{\partial \Tilde{f}}{\partial{z}}(T)$ is well-defined for all $T$, $\frac{\partial \Tilde{f}}{\partial{z}}(T) \neq 0$ and $|\frac{\partial \Tilde{f}}{\partial{z}}(T)| > 0$. If $|\Tilde{\mu}(T)|<1$ for all $T$, then $J_T = |\frac{\partial \Tilde{f}}{\partial z}(T)|^2 (1-|\Tilde{\mu}(T)|^2)>0$ for all $T$. Hence, $\Tilde{f}$ is a local homomorphism. Also, by construction, $\Tilde{f}$ is an identity map on the boundary $\partial \Omega$ and thus $\Tilde{f}|_{\partial \Omega}$ is a homomorphism. 

\begin{figure}
    \centering
    \includegraphics[height=3cm]{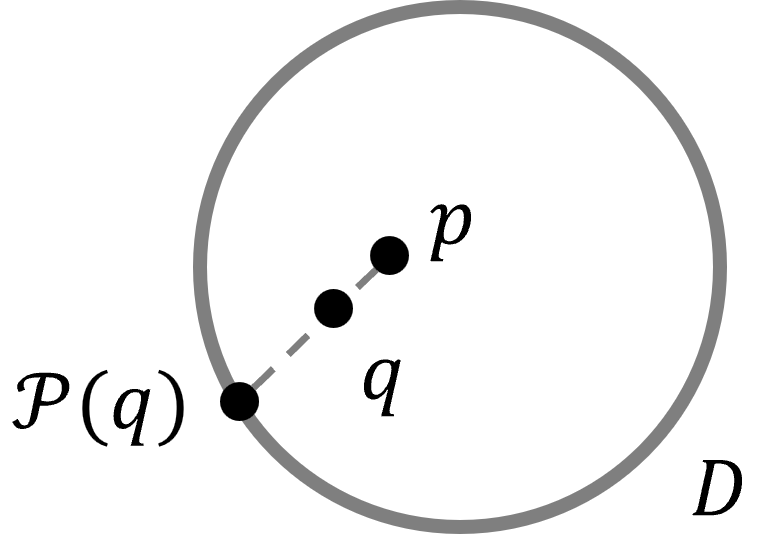}
    \caption{Illustration for Projection construction.}
    \label{fig:projection}
\end{figure}

Let $\phi : \Omega \mapsto D$ be a homeomorphic parameterization of $\Omega$ onto a unit disk $D$. Then, $\hat{f} = \phi \circ \Tilde{f} \circ \phi^{-1} : D\mapsto D$ is a local homeomorphism and $\hat{f}|_{\partial D}$ is homeomorphic. We proceed to show that $\hat{f}$ is a global bijection, which implies $\Tilde{f} = \phi^{-1} \circ \hat{f} \circ \phi$ is a global bijection.

We first prove that $\hat{f}$ is surjective. Suppose not. We can find a point $p$ in the interior of $D$, which is not in the image $\hat{f}(D)$ of $\hat{f}$. Let $\mathcal{P}:D\to \partial D$ be the projection map from $D$ to the unit circle $\partial D$, as depicted in Figure \ref{fig:projection}. Basically, for each $q\in D\setminus\{p\}$, join $p$ and $q$ to get a line. Find the intersection point of the line with $\partial D$ closest to $q$ and define it as $\mathcal{P}(q)$. Then, we have $G = (\hat{f}|_{\partial D})^{-1} \circ \mathcal{P} \circ \hat{f}: D \rightarrow \partial D$ is an identity map on $\partial D$. Note that $(\hat{f}|_{\partial D})^{-1}$ is well-defined since $\hat{f}|_{\partial D}$ is homeomorphic. By the no retraction theorem, such a map does not exist. We conclude that $\hat{f}$ is surjective.

To prove that $\hat{f}$ is injective, it suffices to show that $\hat{f}|_{D\setminus \partial D}:D\setminus \partial D \to D\setminus \partial D$ is injective. Note that $\hat{f}|_{D\setminus \partial D}$ is surjective local homeomorphism and a covering map. Since $D\setminus \partial D$ is simply-connected,  $\hat{f}|_{D\setminus \partial D}$ is a universal covering map. In our case,  the cardinality of the fibers of the universal cover is 1. Hence, $\hat{f}|_{D\setminus \partial D}$ is injective.

\end{proof}

\medskip

As $\Tilde{f}$ is bijective, the final segmentation mask obtained by transforming the template mask by $\Tilde{f}$ is topology-preserving. Note that the Beltrami Adjustment Module is a simple CNN, whose parameters are optimized to minimize $ \mathcal{L}_{trun}$. Alternative speaking, the Beltrami Adjustment Module (BAM) acts as a post-processing step, ensuring the bijectivity of the mapping for each image while preserving the region of interest. If the mapping $f$ from the Deformation Estimation Network (DEN) is already bijective, BAM remains inactive since $\mathcal{L}_{trun}=0$. The Beltrami Solver Network (BS-Net) within BAM remains fixed after pre-training to efficiently solve for the mapping $\Tilde{f}$ from $\Tilde{\mu}$. In the case of 3D, the generalized conformality distortion $K(\Tilde{f})$ in an $n$-dimensional space can be used. This geometric quantity is analogous to measuring quasiconformality in higher dimensions\cite{zhang2022unifying}.

\subsection{Multi-level TPSN}
\begin{figure}[t]
    \centering
    \includegraphics[width = 0.7\textwidth]{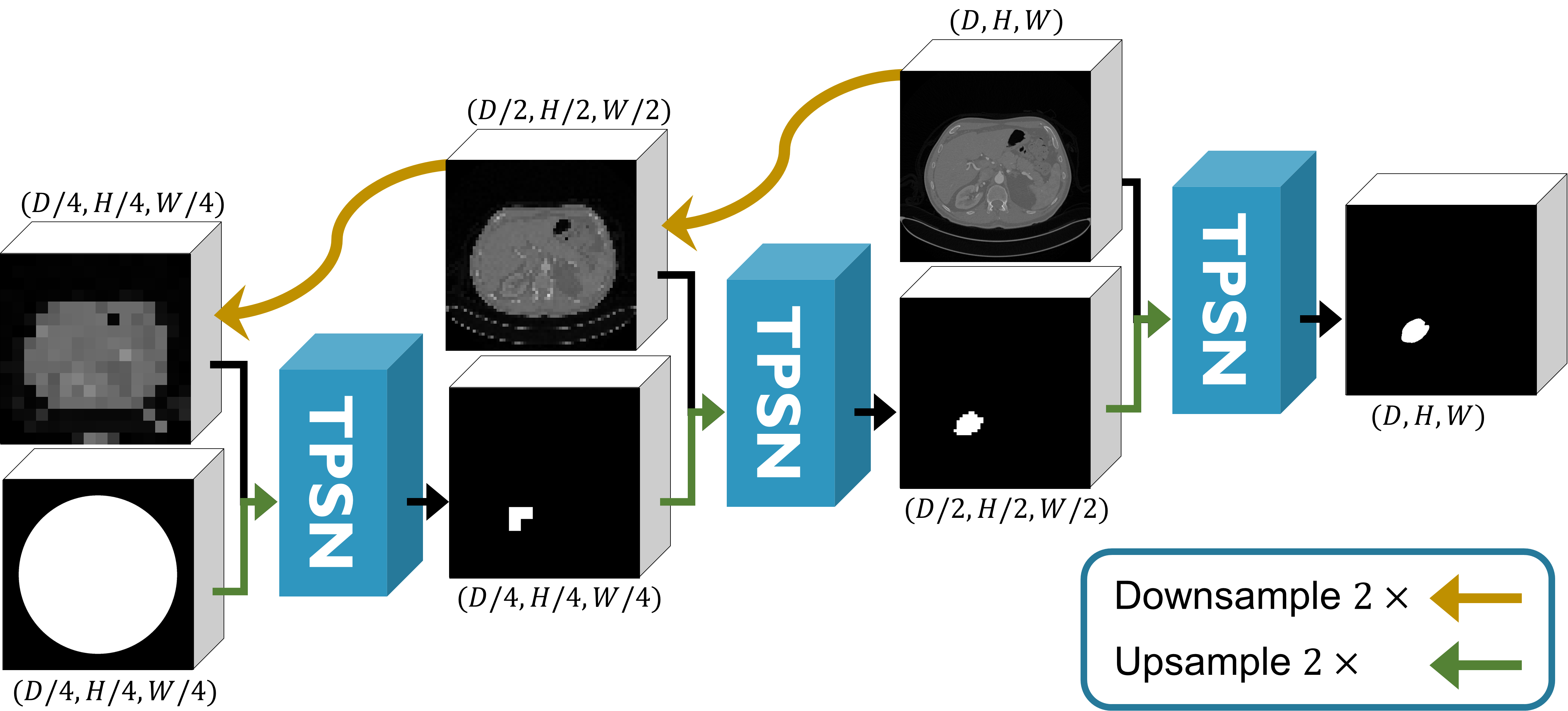}
    \caption{Illustration of the multi-level TPSN. The original image is first downsampled to a low-resolution image, which is then inputted into a TPSN with a prior of the same size. The predicted mask is subsequently upsampled and used as the prior mask in the next layer of the TPSN.}
    \label{fig:mlTPSN}
\end{figure}

To enhance the robustness and accuracy of our proposed model, we utilize a multi-level strategy to construct the "multi-level topology-preserving segmentation network (ml-TPSN)". The pipeline and architecture of ml-TPSN are depicted in Figure \ref{fig:mlTPSN}. We first downsample the original image $I$ into $I^{'}$ and $I^{''}$, with dimensions 1/2 and 1/4 of the original, respectively. $I^{''}$ is then fed into an encoder-decoder network with a coarse template mask of the same size to output a predicted segmentation mask. This coarse predicted mask is upsampled and employed as the template mask $M^{'}$ in the next level for $I^{'}$. $I^{'}$ and $M^{'}$ are then fed into another encoder-decoder network, and the process is repeated. The final encoder-decoder network outputs a mask $M$ with the same resolution as the original image $I$. This multi-level strategy enables our model to extract features in a coarse-to-fine manner, with the predicted mask used as the template mask in the next TPSN layer, leading to a more robust and accurate segmentation, particularly for structures with complex geometry. Additionally, this strategy effectively prevents sub-optimal solutions. The proposed ml-TPSN architecture thus offers improved segmentation results, as demonstrated in our experiments. In our implementation, we utilize the nearest interpolation for both downsampling and upsampling. This choice ensures that the template mask at each level remains binary, containing only integer values.

\subsection{Fill First Dig Second (FFDS) strategy} \label{sec:FFDS} 
\begin{figure}[b]
    \centering
    \includegraphics[width = 0.7\textwidth]{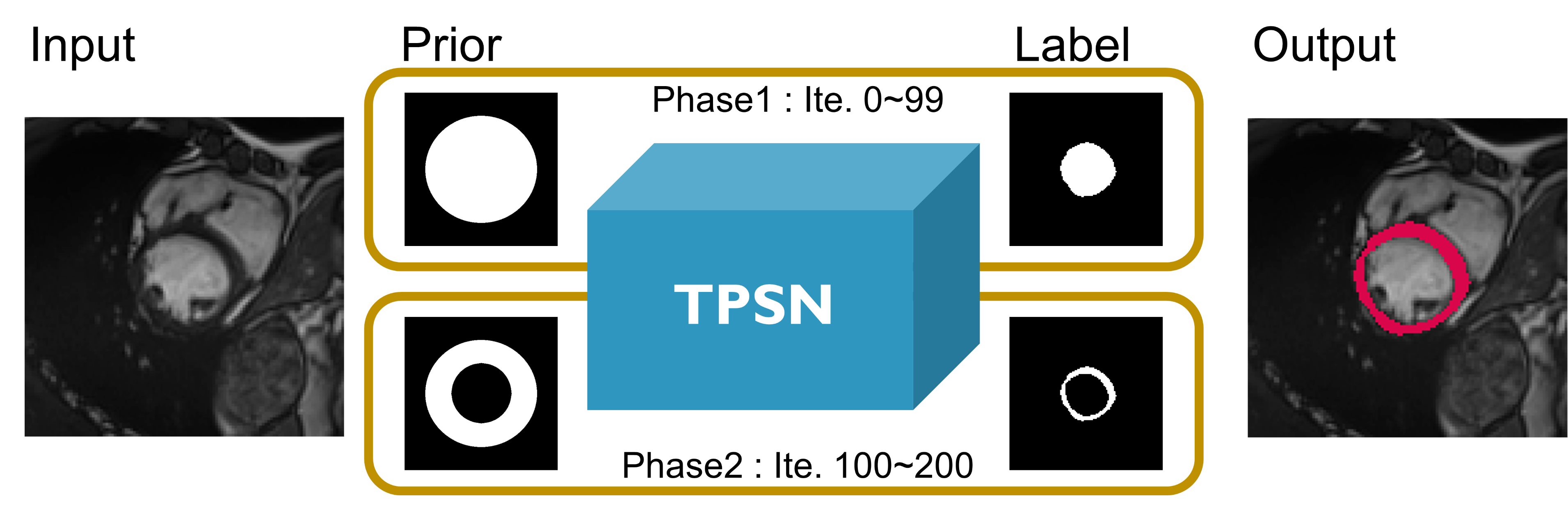}
    \caption{Illustration of Fill First, Dig Second strategy. A disk without holes is deformed to enclose the filled label mask. A hole will be dug out and further optimized to locate the inner boundary.}
    \label{fig:ffds}
\end{figure}

Our proposed framework has demonstrated exceptional efficacy in segmenting multiply-connected objects. To further enhance its performance, we introduce a novel strategy termed Fill First Dig Second (FFDS), depicted in Figure \ref{fig:ffds}. This method involves initially utilizing a simply-connected shape and deforming it to encompass the target region. This process initially fills holes in the label mask, simplifying the segmentation task. During the initial phase, known as the First Fill, the filled template mask is transformed to cover the region of interest whose holes are also filled. Subsequently, in the second phase, holes are excavated from the template to match the prescribed topology of the transformed template mask. The original labeled mask with holes serves as the training label in this phase. Following sufficient training, the transformed mask with holes undergoes further adjustments to ensure its interior edges align with the inner contour of the target region.

The FFDS strategy is crucial. It is important to note that the overlap between the template mask and the label mask plays a crucial role in computing the gradient for training. Without this overlap, the gradient would be null, impeding the training process. By employing the FFDS strategy and filling the holes in the template mask during the first stage, we ensure that there is sufficient overlap between the template mask and the label mask. This allows the model to compute meaningful gradients. As a result, even complex, multiply-connected objects, including those found in other deformable models such as \cite{wyburd2021teds,lee2019tetris}, can be effectively segmented, as shown by our extensive testing.

\subsection{Multi-object Segmentation} Multi-object segmentation is a process that involves the extraction of multiple objects from an image. This technique has found important applications in medical imaging, where the detection and segmentation of multiple anatomical structures can be vital for diagnosis and treatment planning. However, segmenting each individual object using different networks can be both time-consuming and highly costly. While pixel-wise segmentation networks for multi-object segmentation are available in the literature \cite{gibson2018automatic}, the use of deformable models for this task has been relatively unexplored. Our proposed framework can be naturally applied for multi-object segmentation. For single-object segmentation, the output of the encoder-decoder module is typically two or three channels, depending on the dimension of the image. However, for segmenting multiple objects, the number of channels in the output of the TPSN is designed to be $2q$ or $3q$, where $q$ is the number of objects to be segmented. Each group of three channels is then used to give the deformation map necessary to deform the associated template, shown with different colors in the template mask in Figure \ref{fig:framework}. The deformation map is then fed into the Beltrami Adjustment Module to guarantee the bijectivity and is further used to segment the corresponding class of objects in the image.
\section{Discretization of Deformation mappings}
\begin{figure}
    \centering
    \includegraphics[width = 0.5\textwidth]{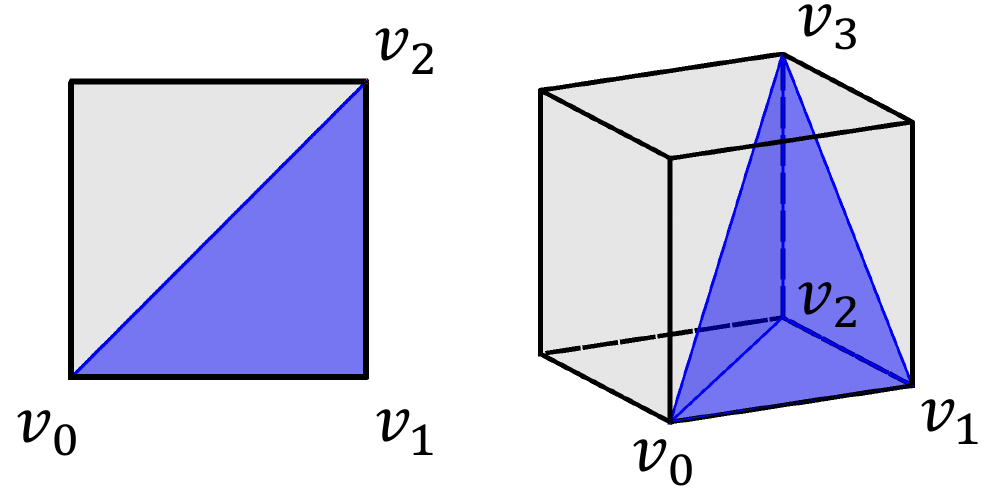}
    \caption{Standard finite element division : Left: for $2$D, a square is divided into two triangles, each element consist of $3$ vertices; Right: for $3$D, a cube is divided into six tetrahedrals, each element consist of $4$ vertices.}
    \label{fig:TetrahedralFEM}
\end{figure}

Our proposed network controls the Jacobian and the Beltrami coefficient of the deformation mapping. In this work, the computation of the Jacobian and Beltrami coefficient is based on the finite element method. The image domain is discretized by a triangulation mesh using the standard finite element division (see Figure~\ref{fig:TetrahedralFEM}). In the 2D case, the image domain is partitioned into small squares and each small square is divided into two triangles. In the 3D case, each cube is divided into six tetrahedrons. The deformation mapping is assumed to be piecewise linear on each triangular face (or each tetrahedron). 

\subsection{Computations of Jacobian and Beltrami coefficients}
We consider the computational of the Jacobian for the 2D case. The computation for the 3D case can be carried out similarly. For an image of size $H \times W$, we parametrize it into a standard domain $[-1,1] \times [-1,1]$. The standard domain is discretized by a regular grid by $G = \{(x_i,y_j)\} = \{(ih_x-1,jh_y-1):i = 0,1 \dots, H-1;j = 0,1 \dots, W-1\}$, where $h_x = \frac{2}{H-1}$ and $h_y = \frac{2}{W-1}$. Each small square in the grid parameterization are divided into $2$ triangles. Since the deformation mapping is considered as a piecewise linear mapping, the Jacobian is a piecewise constant $2\times 2$ matrix on each triangular face. We denote the 3 vertices of a triangle face on the source grid by $v_0$, $v_1$and $v_2$, where $v_i=(x_i,y_i)$ $(i=0,1,2)$. The transformed vertices by the deformation mapping $f=(f_x,f_y)$ are given by $v^*_0$, $v^*_1$ and $v^*_2$, where $v^*_i=(x^*_i,y^*_i)=f(v_i)$ $(i=0,1,2)$. Since $f$ is piecewise linear on each triangular face, we have
\begin{equation}
        \begin{pmatrix}
           x^*\\
           y^*
        \end{pmatrix}
    =   \begin{pmatrix}
           f_{x}(x,y) \\
           f_{y}(x,y)
        \end{pmatrix}
    =   \begin{pmatrix}
           a^x_1 x + a^x_2 y + b^x  \\
           a^y_1 x + a^y_2 y + b^y
        \end{pmatrix}
        \label{eq:matform}
\end{equation}

Hence, the Jacobian of $f$ can be written as
\begin{equation}
    \mathbf{J}(f) = 
    \begin{pmatrix}
    \frac{\partial f_x}{\partial x}   & \frac{\partial f_x}{\partial y}\\
    \frac{\partial f_y}{\partial x}   & \frac{\partial f_y}{\partial y}
    \end{pmatrix} =
    \begin{pmatrix}
           a^x_1 & a^x_2\\
           a^y_1 & a^y_2
    \end{pmatrix}
\end{equation}

Using $v_i^* = f(v_i)$ for $i=0,1,2$, Equation \ref{eq:matform} can be written in the matrix form :
\begin{equation}
\begin{pmatrix}
   x^*_0 & x^*_1 & x^*_2\\
   y^*_0 & y^*_1 & y^*_2
\end{pmatrix}\\
=   \begin{pmatrix}
        a^x_1 & a^x_2& b^x \\
        a^y_1 & a^y_2 & b^y \\
    \end{pmatrix}
    \begin{pmatrix}
       x_0 & x_1 & x_2\\
       y_0 & y_1 & y_2\\
       1   & 1   & 1
    \end{pmatrix}.
\end{equation}

It can be further reduced into 
\begin{equation}
\begin{split}
        \begin{pmatrix}
           x^*_0-x^*_2 & x^*_1-x^*_2\\
           y^*_0-y^*_2 & y^*_1-y^*_2
        \end{pmatrix}
    =   \begin{pmatrix}
            a^x_1 & a^x_2\\
            a^y_1 & a^y_2\\
        \end{pmatrix}
        \begin{pmatrix}
           x_0-x_2 & x_1-x_2\\
           y_0-y_2 & y_1-y_2
        \end{pmatrix}
        \label{eq:reduce}
\end{split}
\end{equation}

By writing Equation \ref{eq:reduce} as $X = \mathbf{J} Y$, the Jacobian of $f$ can be obtained by $\mathbf{J} = XY^{-1}$. The determinant of the Jacobian matrix can then be computed, which is a constant on each triangular face. 

With the above discretization, the Beltrami coefficient of $f$ can also be computed. Note that the Beltrami coefficient is defined by the first derivatives of $f$, which are constants on each triangular face. As such, the Beltrami coefficient is a constant complex number on each triangular face. More specifically, the Beltrami coefficient can be computed by
\begin{equation}
\begin{split}
    \mu & = \frac{(\frac{\partial f_x}{\partial x}- \frac{\partial f_y}{\partial y}) + i(\frac{\partial f_y}{\partial x}+ \frac{\partial f_x}{\partial y})}{(\frac{\partial f_x}{\partial x}+ \frac{\partial f_y}{\partial y}) + i(\frac{\partial f_y}{\partial x}- \frac{\partial f_x}{\partial y})} \\
    &= \frac{(a_1^x-a_2^y)+i(a_1^y+a_2^x)}{(a_1^x+a_2^y)+i(a_1^y-a_2^x)}
    \label{eq:bc}
\end{split}
\end{equation}

\begin{figure}
    \centering
    \includegraphics[width = 0.7\textwidth]{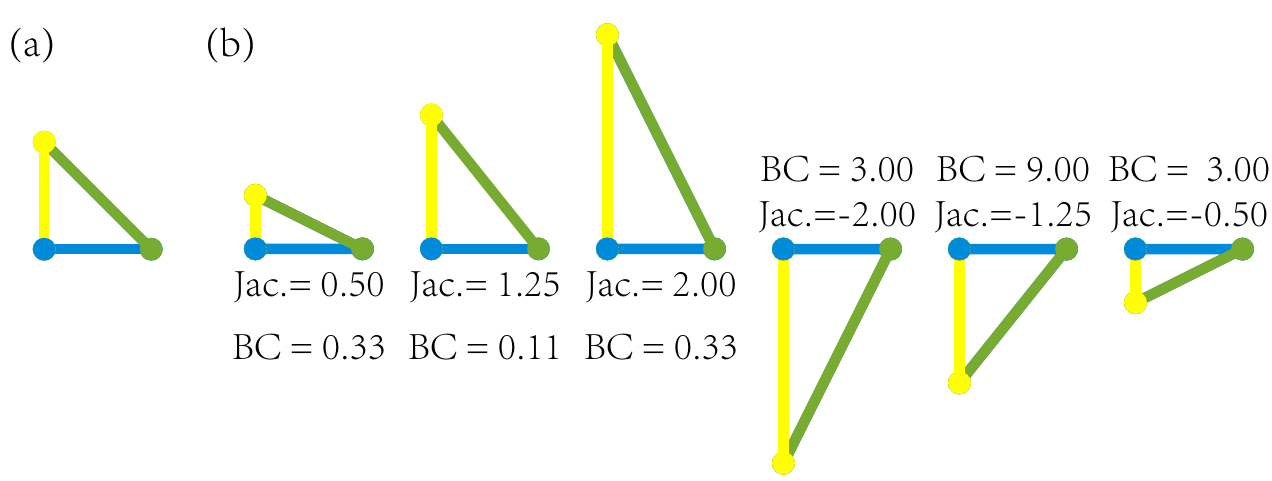}
    \caption{Illustration of deformations with different values of Jacobian and the norm of Beltrami Coefficients. A positive Jacobian determinant (or a Beltrami coefficient whose norm is less than $1$) represent an orientation-preserving deformation. A negative Jacobian determinant (or a Beltrami coefficient whose norm is greater than $1$) represents an orientation change in the mapping.}
    \label{fig:JacobianTriangle}
\end{figure}

The Jacobian determinant and the Beltrami coefficient provides information about the change of orientation on each triangular face under $f$. Figure \ref{fig:JacobianTriangle} illustrates how the orientation of a map can be described by the Jacobian determinant. A triangle with the orientation "Green-Blue-Yellow" clockwise in (a) is transformed into the first three triangles in (b), which are all orientation-preserving. In these cases, the Jacobian determinants of the deformation maps are all positive and the magnitudes of the Beltrami coefficients are all less than 1. The triangle in (a) is further transformed into the last three triangles, which are orientation-reversing. In these cases, the Jacobian determinants of deformation maps are all negative and the magnitudes of the Beltrami coefficients are all greater than 1.

\subsection{Reconstruction of quasiconformal mappings}
Given a piecewise constant Beltrami coefficient $\mu$, the associated piecewise linear mapping $f$ can be reconstructed by solving a linear system, called the Linear Beltrami Solver \cite{lui2013texture}. Given a triangular face $T$, we denote $\mu(T)$ by $\mu(T) = \mu_T = \rho_T + \sqrt{-1}\tau_T$. On each $T$, the Beltrami's equation can be written in a matrix form:
\begin{equation}
    \begin{pmatrix}
    -a_2^y\\
    a_1^y
\end{pmatrix} = \frac{1}{1-\rho_T^2 - \tau_T^2}\begin{pmatrix}
    \alpha_1(T) & \alpha_2(T)\\
    \alpha_2(T) & \alpha_3(T)
\end{pmatrix}\begin{pmatrix}
    a_1^x\\
    a_2^x
\end{pmatrix},
\end{equation}
\noindent where $\alpha_1(T) = -(1-\rho_T)^2 - \tau_T^2$, $\alpha_2(T) = -2\tau_T$ and $\alpha_3(T) = -(1+\rho_T)^2 - \tau_T^2$. Applying the discrete divergence $Div$ on both sides, we obtain a linear system
\begin{equation}\label{LBS}
    A_{\mu} {\bf f} = {\bf b},
\end{equation}
\noindent where ${\bf f}$ is a vector whose $i$-th entry is given by $f(v_i)$ of vertices $v_i$, $A_{\mu}$ is a matrix depending on $\mu$ and ${\bf b}$ is a vector depends on $\mu$ and the boundary conditions. As such, $f$ can be reconstructed from $\mu$ by solving Equation~\ref{LBS}. The representation of $f$ by $\mu$ is beneficial since $\mu$ measures the local geometric distortion under $f$. We can thus easily control the regularlity of $f$ by controlling $\mu$.

To facilitate the optimization process in the Beltrami Adjustment Module, we apply the BS-Net \cite{chen2021deep}, which is a neural network to output $f$ from an input $\mu$. Given a pre-trained and frozen BS-Net, the backward propagation can then be carried out effectively during the training process.

\section{Experiment}

In this section, we will evaluate the efficiency of our proposed models through various experiments. To systematically examine the impact of different parameters and the influence of the template mask shape, we will conduct some self-ablation studies. Our proposed segmentation model will be tested on multiple real image datasets. Moreover, we will compare our methods with other state-of-the-art topology-preserving learning-based models such as TEDSNet\cite{wyburd2021teds} and TeTrIs\cite{lee2019tetris}. Additionally, we will compare our models with other pixel-wise segmentation networks, such as the baseline U-Net model \cite{ronneberger2015u} and the double U-Net model \cite{jha2020doubleu}. Finally, we will report quantitative measurements of the segmentation results. 

To ensure a fair comparison, all experiments are conducted on the same platform and under identical training settings, including batch number and running epochs. The backbone networks used for each method have the same number of layers to minimize the impact of other factors. In addition, all methods employ Dice loss as the fidelity term for supervised training, except for TopoLoss \cite{hu2019topology} and BettiMatching \cite{stucki2023topologically}, which introduced their own fidelity loss functions. No supplementary data preprocessing or postprocessing is employed for any of the methods, facilitating a more transparent evaluation of the progress achieved by each approach. The objective of these experiments is not to attain a higher score compared to state-of-the-art methods through data manipulation but rather to assess our methods' capacity to preserve topology in segmentation.

\noindent\textbf{Dataset} In this work, we utilized four datasets to evaluate the effectiveness of our proposed models. The 3D volumetric dataset of kidney images \cite{heller2021state} was used for the self-ablation study. Out of the $300$ images in the dataset, $210$ were used for training, and the remaining $90$ were used for testing. All images were central-cropped and resized to $128\times 128 \times 64$. To evaluate the unsupervised TPSN model for segmenting skin lesions\cite{tschandl2018ham10000,tschandl2020human}, we used the Ham10000 dataset, considering only the simply-connected lesions. We removed the images containing disconnected lesion regions in the original dataset, and the final dataset comprised $9981$ images. Out of these, $8981$ were used for training, and the remaining $1000$ images were used for testing. All images were resized to $256\times 256$. To evaluate our model for segmenting genus-1 structures, we used the ACDC \cite{bernard2018deep} dataset, where $5$ slices containing circular vessel structures from $100$ cases were extracted and used as the dataset. Out of these, $90$ cases comprising $450$ images were used for training, while the remaining $10$ cases ($50$ images) were used for testing. We used the KiTS21\cite{heller2021state} dataset for 3D segmentation of simply-connected structures, following the same process as in the self-ablation study. Finally, we evaluated the capability of our methods for multi-object segmentation on BTCV dataset\cite{btcv2015}. We divided the whole $30$ cases in the dataset into $25$ and $5$ for training and testing, respectively, and used only $8$ organs (spleen, right kidney, left kidney, gallbladder, liver, stomach, right adrenal gland, left adrenal gland) in this work. All images were centrally cropped and resized to $128 \times 128 \times 128$. The intensity of all images was normalized to $[0,1]$.

\noindent\textbf{Computing resources and Parameters}
All models were trained for $300$ epochs using RMSprop optimizer with a learning rate of $1\times 10^{-5}$. For experiments on KiTS21 and BTCV, a batch size of $8$ was used, while a batch size of $16$ was used for experiments on ACDC. The models were trained on a CentOS 8.1 central cluster computing node with two Intel Xeon Gold 5220R 24-core CPUs and one NVIDIA A100 Tensor Core GPUs. In the supervised segmentation, the weighting parameters $\lambda_{Jac}$ and $\lambda_{Lap}$ were set to be $1.0$ and $0.01$, respectively, unless otherwise specified.

\noindent\textbf{Metrics}
The following metrics will be used for quantitative measurements. Denote the domain in the predicted mask and in the label mask by $X$ and $Y$, respectively. \begin{itemize}
    \item \textit{Dice score} is defined as $DSC=\frac{2 |X \cap Y|}{|X|+|Y|}$. The unit for Dice in the tables are $\%$.
    \item \textit{Hausdorff distance} is defined as $d_H (X,Y) = max\{\sup\limits_{x \in X} d(x,Y) ,  \sup\limits_{y \in Y} d(X,y)\}$. 
    \item \textit{Betti error} used in this paper is defined as the $0$-dimensional Betti number \cite{hu2019topology, clough2020topological}. We use $8$-connected component labeling algorithm to count for the regions.
    \item $\mathcal{L}_{Jac}$ reported in the tables is the value of $ReLU$-Jacobian loss computed as Equation \ref{eq:relujac}. It is zero for any bijective mappings.
    \item \textit{BettiMatching Error} is a topologically and feature-wise accurate metric proposed in \cite{stucki2023topologically}.
\end{itemize}

\begin{figure*}[t]
    \centering
    \includegraphics[width = \textwidth]{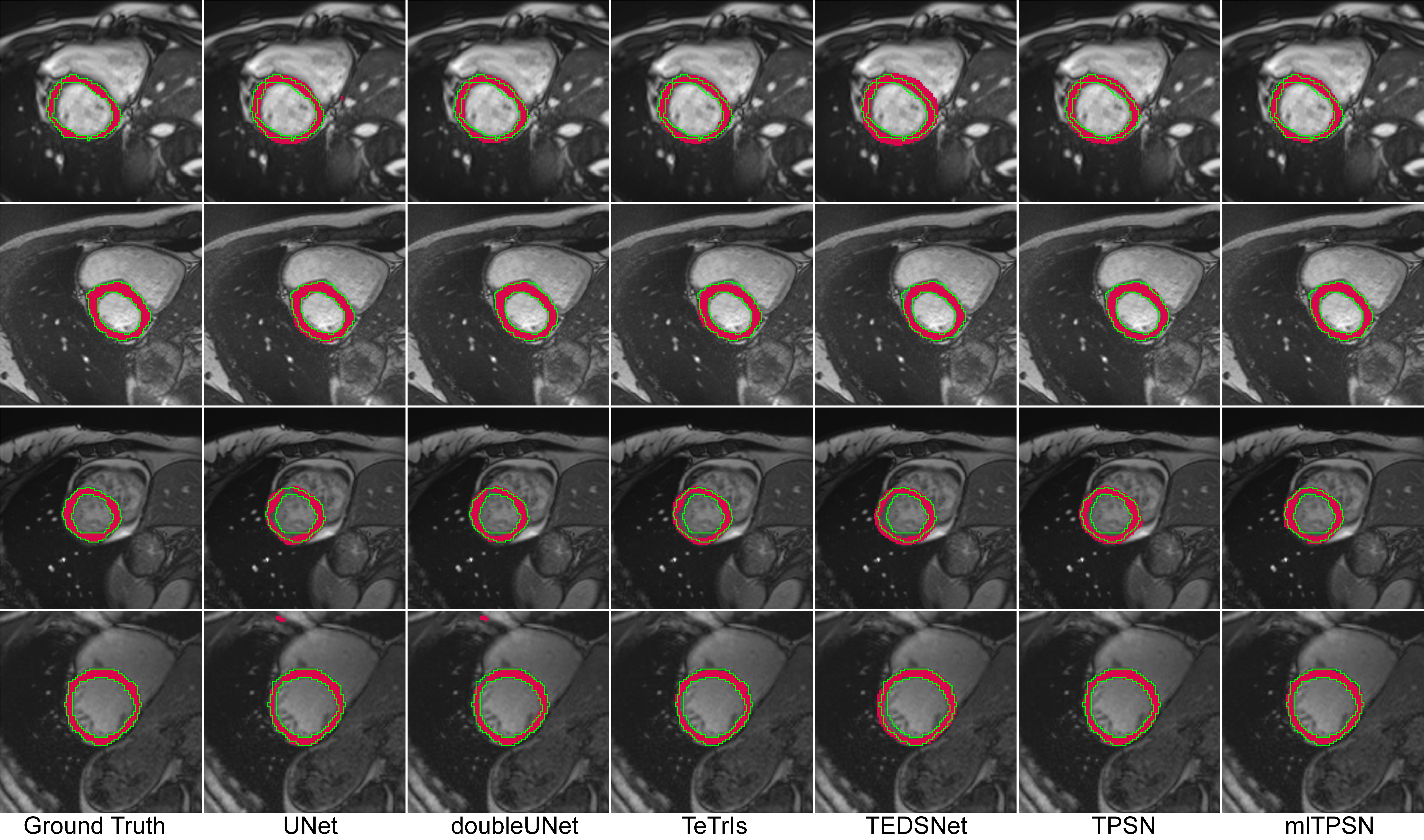}
    \caption{Segmentation results comparing different methods for the ACDC dataset. Conventional pixel-wise approaches often produce isolated noisy regions (indicated by yellow arrow) and discontinuity errors (indicated by blue arrow). Other state-of-the-art deformable models may result in suboptimal outcomes due to over-constraint.}
    \label{fig:ACDC}
\end{figure*}

\subsection{Ablation Studies on DEN}
We conducted ablation studies to explore the impact of various parameters and evaluate the influence of the template mask in the DEN module of the proposed model. All the networks use a single-level setting except the experiment evaluating the influence by the number of levels. The template mask for experiments on $\lambda_{Jac}$, $\lambda_{Lap}$ and No. of levels are the mean shape over all the label masks in the training set calculated by \cite{lin2022harmonic}. The weighting parameters $\lambda_{Jac}$ and $\lambda_{Lap}$ were set to be $1.0$ and $0.01$ unless otherwise specified.

\newcolumntype{C}{>{\centering\arraybackslash}p{0.08\columnwidth}}
\begin{table}
\small
\centering
\begin{tabular}{c|p{0.15\columnwidth}|C|C|C|C|C}
\toprule
Parameter                           & Value     & Dice          & HD            & Betti & $\mathcal{L}_{Jac}$   & $\|\mu\|$\\\hline
\multirow{6}{*}{$\lambda_{Jac}$}   & $0$       & \B{93.01}     & 4.34          & 1.27  & 27.34                 &159.56\\
                                    & $10^{-2}$ & 92.96         & 4.12          & 1.08  & 0.25                  &12.47\\
                                    & $10^{-1}$ & 92.87         & 4.08          & 1.03  & 0.03                  &3.82\\
                                    & $1.0$     & 92.65         & 3.86          & \B{1.00} & 0.00               &0.48\\
                                    & $10^1$    & 92.88         & \B{3.84}      & \B{1.00} & 0.00               &0.43\\
                                    & $10^2$    & 89.11         & 4.41          & \B{1.00} & 0.00               &0.55\\\hline
\multirow{6}{*}{$\lambda_{Lap}$}   & $0$       & \B{92.72}     & 3.97          & 1.03  & 0.07                 &4.65\\
                                    & $10^{-4}$ & 92.69         & 3.92          & \B{1.00} & 0.03               &1.58\\
                                    & $10^{-2}$ & 92.65         & \B{3.86}      & \B{1.00} & 0.00               &0.48\\
                                    & $1.0$     & 92.60         & 3.89          & \B{1.00} & 0.00              &0.43\\
                                    & $10^2$    & 92.63         & 3.87          & \B{1.00} & 0.00              &0.47\\
                                    & $10^4$    & 90.87         & 4.13          & \B{1.00} & 0.00              &0.38\\\hline
                                    & 1         & 92.65         & 3.86          & \B{1.00} & 0.00               &0.48\\
No. of                              & 2         & 93.21         & 3.82          & \B{1.00} & 0.00               &0.61\\
levels                              & 3         & 93.63         & \B{3.79}      & \B{1.00} & 0.00               &0.59\\
                                    & 4         & \B{93.71}     & \B{3.79}      & \B{1.00} & 0.00               &0.43\\\hline  
\multirow{4}{*}{Shape}              & Disk      & 92.49         & 3.91          & 1.00  & 0.00                  &0.57\\
                                    &Ellipse    & 92.51         & 3.90          & 1.00  & 0.00                  &0.42\\
                                    &Sqaure     & 92.38         & 3.98          & 1.00  & 0.00                  &0.48\\
                                    &Mean       & \B{92.65}     & \B{3.86}      & 1.00  & 0.00                  &0.48\\\hline
\multirow{2}{*}{Input}              &Only $I$   & 92.63         & 3.88          & 1.00  & 0.00                  &0.43\\
                            &$I$ and $M_{temp}$ & \B{92.65}     & \B{3.86}      & 1.00  & 0.00                  &0.48\\\hline
\bottomrule
\bottomrule
\end{tabular}
\captionof{table}{Quantitative measurements for the self-ablation studies.}
\label{tb:self}
\end{table}
\begin{figure}
    \centering
    \includegraphics[width = 0.5\textwidth]{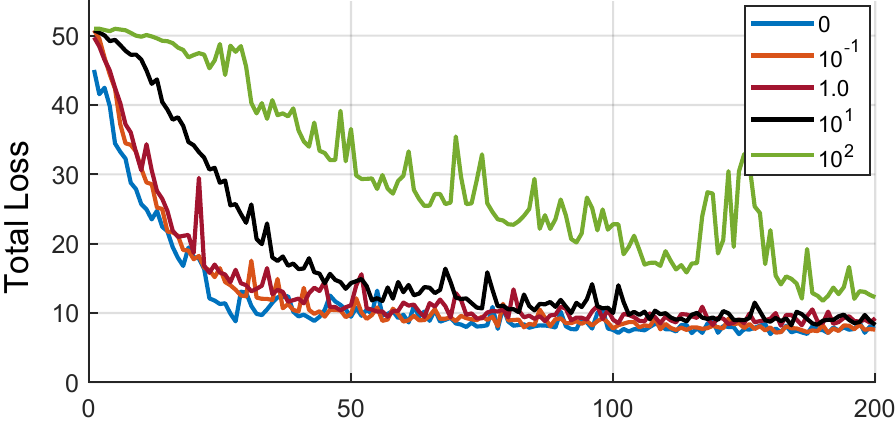}
    \caption{The convergence analysis was performed using different values of $\lambda_{Jac}$. The results demonstrate that larger values of $\lambda_{Jac}$ slow down the training process, but the convergence is still acceptable, typically within 100 epochs. However, when $\lambda_{Jac}$ exceeds 10, the segmentation results become less accurate due to over-constraint.}
    \label{fig:lambdagraph}
\end{figure}

\smallskip

\noindent\textbf{Influence of $\lambda_{Jac}$ on DEN:}
We begin by examining the significance of our proposed Jacobian regularization term, denoted as $\mathcal{L}_{Jac}$. As all mapping adjusted by BAM will be bijective, we conducted experiments with varying weighting parameters $\lambda_{Jac}$ only on the output of DEN. In this way, the impact of various values of $\mathcal{L}_{Jac}$ on the bijectivity of the outputted mapping can be assessed. Specifically, we used the following parameters: $\lambda_{Jac}=0,10^{-2},10^{-1},1,10,10^2$. Table \ref{tb:self} reports the quantitative measurements of the segmentation results using different metrics, averaging the results from $5$ independent experiments. The table shows that the segmentation results achieved without using the ReLU Jacobian regularization yield the highest Dice score but the worst Hausdorff distance due to topological errors in the predicted mask, reflected in the Betti number. However, with $\lambda_{Jac}=1.0$, we can achieve topological correctness without sacrificing the Dice score too much. As reported in the ReLU-Jacobian error, we observe that choosing $\lambda_{Jac}\geq 1.0$ is sufficient to ensure the Jacobian determinant remains positive and bijective. If $\lambda_{Jac}$ exceeds $10$, the segmentation results become less accurate due to over-constraint. Figure \ref{fig:lambdagraph} illustrates the convergence of the loss function for each $\lambda_{Jac}$, showing that larger weighting slows down the convergence. Thus, to ensure the topology-preserving while maintaining accurate segmentation and fast convergence, we choose $\lambda_{Jac}=1.0$ hereafter.

\smallskip

\noindent\textbf{Influence of $\lambda_{Lap}$ on DEN:} When $\lambda_{Lap}$ is less than $10^{-2}$, the mapping is not bijective as reflected by the non-zero value of $\mathcal{L}_{Jac}$ and the supermum of $\mu$ being greater than $1$. The weighting parameter can vary widely according to Table \ref{tb:self}. However, when $\lambda_{Lap}=10^{4}$, the problem becomes over-constrained, resulting in significant degradation of the segmentation result as indicated by the Dice score.

\smallskip

\noindent\textbf{Influence of levels:} We evaluate the impact of the number of levels used in the multi-level TPSN by setting $\lambda_{Jac}=1.0$ and $\lambda_{Lap}=10^{-2}$. The segmentation results are reported in TABLE \ref{tb:self}, and it is evident that the mlTPSN with varying numbers of levels can yield segmentation results with correct topology. The results show improvement from $1$ to $3$ levels, with only slight differences between $3$ and $4$ levels. Thus, in the after experiments, we use a 3-level configuration for mlTPSN.

\smallskip

\noindent\textbf{Influence of template mask} We also explore the robustness of our model under various template masks. To this end, we conducted experiments with four different template shapes: (i) a disk, (ii) an ellipse, (iii) a square, and (iv) the average shape from a dataset. We selected the Ham10000 dataset for this study and the quantitative results are reported in TABLE \ref{tb:self}. The results indicate that the segmentation outcomes obtained using different template synthetic shapes (i.e., shapes (i), (ii), and (iii)) are similar. However, the segmentation results achieved using the mean shape are the best, demonstrating that a template shape closer to the ground truth can enhance segmentation accuracy, as expected.

\smallskip

\noindent\textbf{Influence of Inputs} We conducted ablation studies on the input, specifically exploring scenarios where only the image or both the image and template were used as input. When using the same template mask as input, we excluded the direct input of the template information into the network. Even without the input of the template information, the network learned to warp the template into the region of interest during training. Therefore, whether we explicitly provided the template mask as an input to the network or not did not make a difference, as evidenced by the results shown in Table \ref{tb:self}.

\subsection{Segmentation of real images}

In this subsection, we evaluate the performance of our proposed models for segmenting real images, including both supervised and unsupervised image segmentation.

\subsubsection{Supervised segmentation} 

We evaluated the efficiency of our proposed model for supervised segmentation by utilizing two image datasets, namely the ACDC2D and KiTS21, both of which have labeled segmentation masks. Additionally, we conduct a thorough comparison of our segmentation results with those of other state-of-the-art methods.

\newcolumntype{D}{>{\centering\arraybackslash}p{0.08\columnwidth}}
\begin{table}
\small
\centering
\begin{tabular}{p{0.2\columnwidth}|D|D|D|c|D}
\toprule
Method      & Dice      & HD        & Betti     & BettiMatching & $\mathcal{L}_{Jac}$\\\hline
UNet        & 84.31     & 2.46      & 1.08      & 2.28          & $\setminus$        \\ 
wUNet       & \B{86.10} & 2.26      & 1.04      & 2.22          & $\setminus$        \\ \cline{1-6}
TopoLoss    & 82.85     & 2.73      & 1.02      & 2.14          & $\setminus$        \\ 
BetMatCH    & 82.68     & 2.66      & \B{1.00}  & 2.10          & $\setminus$        \\ \cline{1-6}
TeTrIs      & 79.86     & 5.33      & 1.14      & 2.54          & 7.49               \\
TEDS        & 78.90     & 4.80      & \B{1.00}  & 2.28          & 1.27               \\\cline{1-6} 
TeTrIs(FFSD)& 80.26     & 3.54      & 1.04      & 2.15          & 2.49               \\
TEDS(FFSD)  & 81.54     & 3.47      & \B{1.00}  & 2.11          & 1.46               \\\cline{1-6} 
TPSN        & 82.75     & 2.37      & \B{1.00}  & 2.06          & 0.00               \\ 
mlTPSN      & 85.15     & \B{2.04}  & \B{1.00}  & \B{2.02}      & 0.00               \\\hline
\bottomrule
\end{tabular}
\captionof{table}{Quantitative comparison with other methods for 2D double-connected component segmentation on the ACDC2D dataset.}
\label{tb:acdc}
\end{table}
                        
\begin{table}
\small
\centering
\begin{tabular}{p{0.2\columnwidth}|D|D|D|D}
\toprule
Method      & Dice      & HD        & Betti     & $\mathcal{L}_{Jac}$\\\hline
UNet      & 93.41     & 4.92      & 1.68      & $\setminus$   \\ 
wUNet     & \B{94.26} & 4.48      & 1.31      & $\setminus$   \\ \cline{1-5}
TeTrIs    & 91.03     & 4.44      & 1.07      & 0.20          \\
TEDS   & 90.07     & 4.27      & \B{1.00}  & 0.00          \\\cline{1-5}
TPSN      & 92.65     & 3.86      & \B{1.00}  & 0.00          \\ 
mlTPSN    & 93.63     & \B{3.80}  & \B{1.00}  & 0.00          \\\hline
\bottomrule
\end{tabular}
\captionof{table}{Quantitative comparison with other methods for 3D single-connected component segmentation on the KiTS21 dataset.}
\label{tb:kits}
\end{table}

\textbf{2D image segmentation} 
We applied our proposed segmentation framework to the ACDC2D dataset to extract doubly-connected structures from 2D images. We compare our results with several state-of-the-art methods, including two non-deformation-based baseline models: UNet \cite{ronneberger2015u} and doubleUnet \cite{jha2020doubleu}; two methods utilizing their proposed topology losses: TopoLoss\cite{hu2019topology} and BettiMatching \cite{stucki2023topologically}; as well as two learning-based deformation-based models: TEDSNet \cite{wyburd2021teds} and TeTrIs \cite{lee2019tetris}. The template mask is the mean shape over all the label masks in the training set calculated by \cite{lin2022harmonic}.

Besides those methods, the FFDS strategy described in Section \ref{sec:FFDS} is applied to all deformable approaches in the comparison. In particular, the first phase is executed during the initial $50$ epochs to delineate the outer boundary, while the remaining epochs focus on approximating the inner boundary. Additionally, we report results obtained without employing the FFDS strategy to underscore its advantages.

Figure \ref{fig:ACDC} illustrates the segmentation results obtained using different methods. It is evident that methods lacking topology regularization, such as UNet and doubleUnet, often produce incorrect isolated components (highlighted by the yellow arrow) that are located far away from the object to be segmented. In contrast, the utilization of TopoLoss \cite{hu2019topology} and BettiMatching \cite{stucki2023topologically} demonstrates better topology control. However, their regularization, implemented as soft constraints, can still result in incorrect topology. TeTrIs \cite{lee2019tetris}, which performs segmentation by generating a mapping to warp a template mask, fails to preserve topology as the predicted mapping with a small Laplacian does not indicate bijectivity. Although TEDSNet successfully preserves topology, the generated mapping may be too restrictive to accurately deform the template to enclose the region of interest. Our methods, depicted in the last two columns, consistently yield topology-preserved segmentation results.

These results are further supported by the quantitative findings reported in Table \ref{tb:acdc}. Both TPSN and multi-level TPSN consistently produce results with correct topology, as indicated by the predicted mapping possessing a positive Jacobian (as reflected by the null value of the ReLU-Jacobian error). Without noisy outliers, our method outperforms others in terms of Hausdorff distance. In terms of the Dice score, our method achieves comparable results to pixel-wise classification methods. Furthermore, our methods outperform others in terms of BettiMatching error \cite{stucki2023topologically}, highlighting the advancement of our approach.

\begin{figure}[h]
    \centering
    \includegraphics[width = 0.7\textwidth]{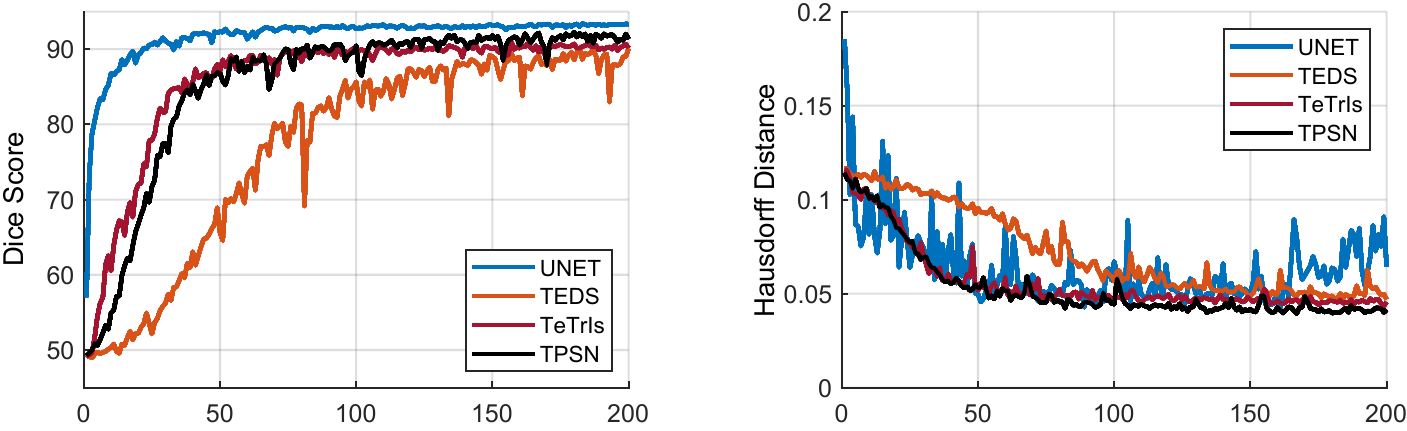}
    \caption{The convergence analysis of various methods reveals that UNet outperforms the rest in terms of the Dice score and speed of convergence. This is due to its flexibility to segment through pixel-wise classification. However, when using the Hausdorff distance metric, UNet is the worst due to insufficient geometric regularization. On the other hand, our methods exhibit superior performance by Hausdorff distance and achieve comparable results to UNet in terms of Dice score.}
    \label{fig:KiTS21}
\end{figure}

\textbf{3D image segmentation} 
We evaluated our segmentation framework on the KiTS21 dataset to segment organs from 3D volumetric images. Specifically, we focused on segmenting the kidney, a simply-connected organ with a single component, using a 3D simply-connected cube as the template mask. In this experiment, only the left kidney with labels was utilized, and the quantitative results are detailed in Table \ref{tb:kits}.

We observed that pixel-wise segmentation methods generally achieve higher Dice scores. This can be attributed to the inherent flexibility of pixel-wise classification approaches. However, these methods often suffer from topological errors and produce noisy isolated regions, resulting in higher Hausdorff distances. Among deformable approaches, TeTrIs performs relatively well in terms of Hausdorff distance but struggles to preserve the topology of the prior template mask. On the other hand, TEDSNet excels in topology preservation but exhibits significantly lower Dice scores and Hausdorff distances, possibly due to over-constrained conditions.

In contrast, our method capitalizes on the bijectivity of the predicted mapping for warping the template masks. This is evident from the zero value of $\mathcal{L}_{Jac}$, indicating that our method is free from topological errors. Moreover, our method achieves a significant improvement in terms of Hausdorff distance, while still obtaining a comparable Dice score to that of pixel-wise classification methods.

\begin{figure}[hb!]
    \centering
    \includegraphics[width = 0.7\textwidth]{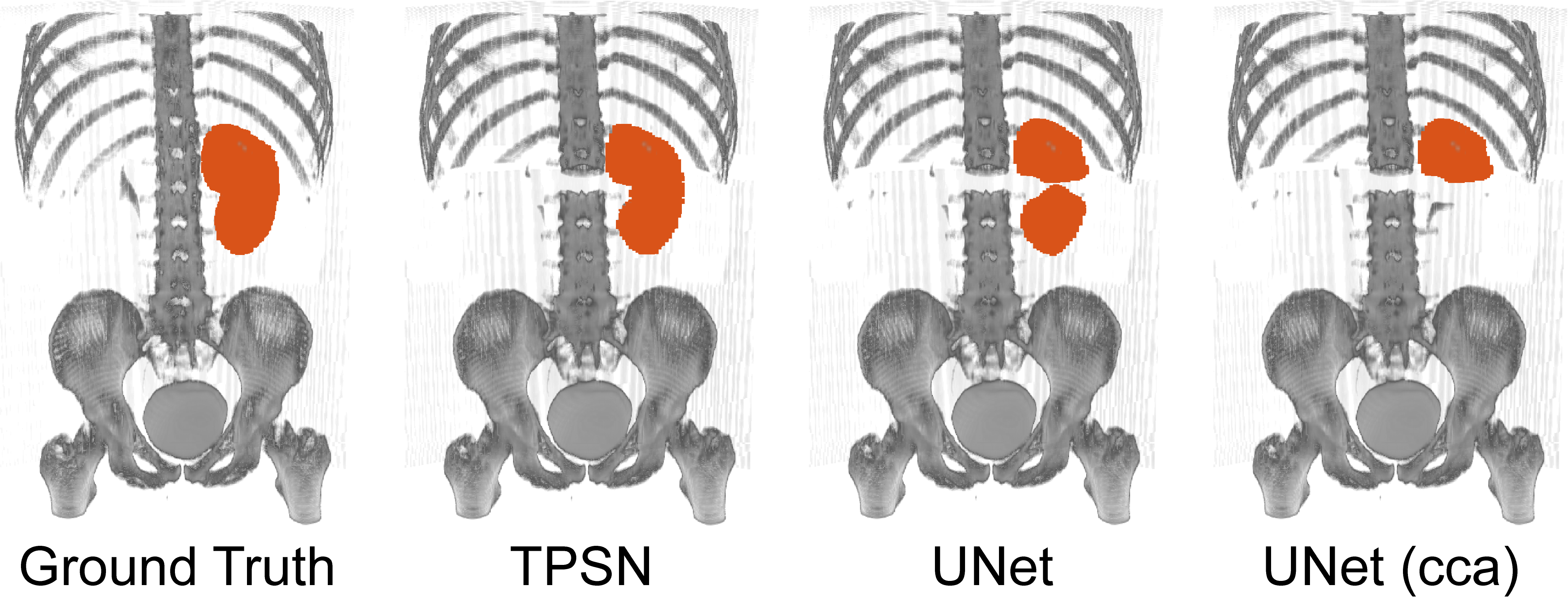}
    \caption{3D segmentation results of the KiTS21 dataset with missing slices due to information loss. Pixel-wise segmentation methods like UNet produce two separate parts, as shown in the image. This issue cannot be resolved using connected component analysis, as demonstrated.}
    \label{fig:interruption}
\end{figure}

The convergence of different models in terms of different metrics was also analyzed, as shown in Figure \ref{fig:KiTS21}. The pixel-wise segmentation model shows faster convergence, while Dice loss (which is $1 - $\textit{Dice Score}) converges the best due to the flexibility provided by pixel-wise classification. Among deformable methods, our model achieves the best convergence regarding Dice loss, while TEDSNet has the slowest convergence among the four methods. In terms of Hausdorff distance, we observe that all deformable segmentation models converge in a similar gradual pattern, while the pixel-wise segmentation model shows an unstable, non-convergent curve.

We also studied the tolerability of our model to information loss after sufficient training. We assumed that some slices of the scan were missing, resulting in missing information. The incomplete scans were then fitted into the trained network without processing. The segmentation results obtained by different methods are shown in Figure \ref{fig:interruption}. Our method produces an accurate segmentation result with the correct topology even with missing information. In contrast, the segmentation result by UNet has two separate components, and using connected component analysis to correct the topology in such a case cannot solve the issue.

\subsubsection{Unsupervised segmentation}
\begin{figure}
    \centering
    \includegraphics[width = 0.6\textwidth]{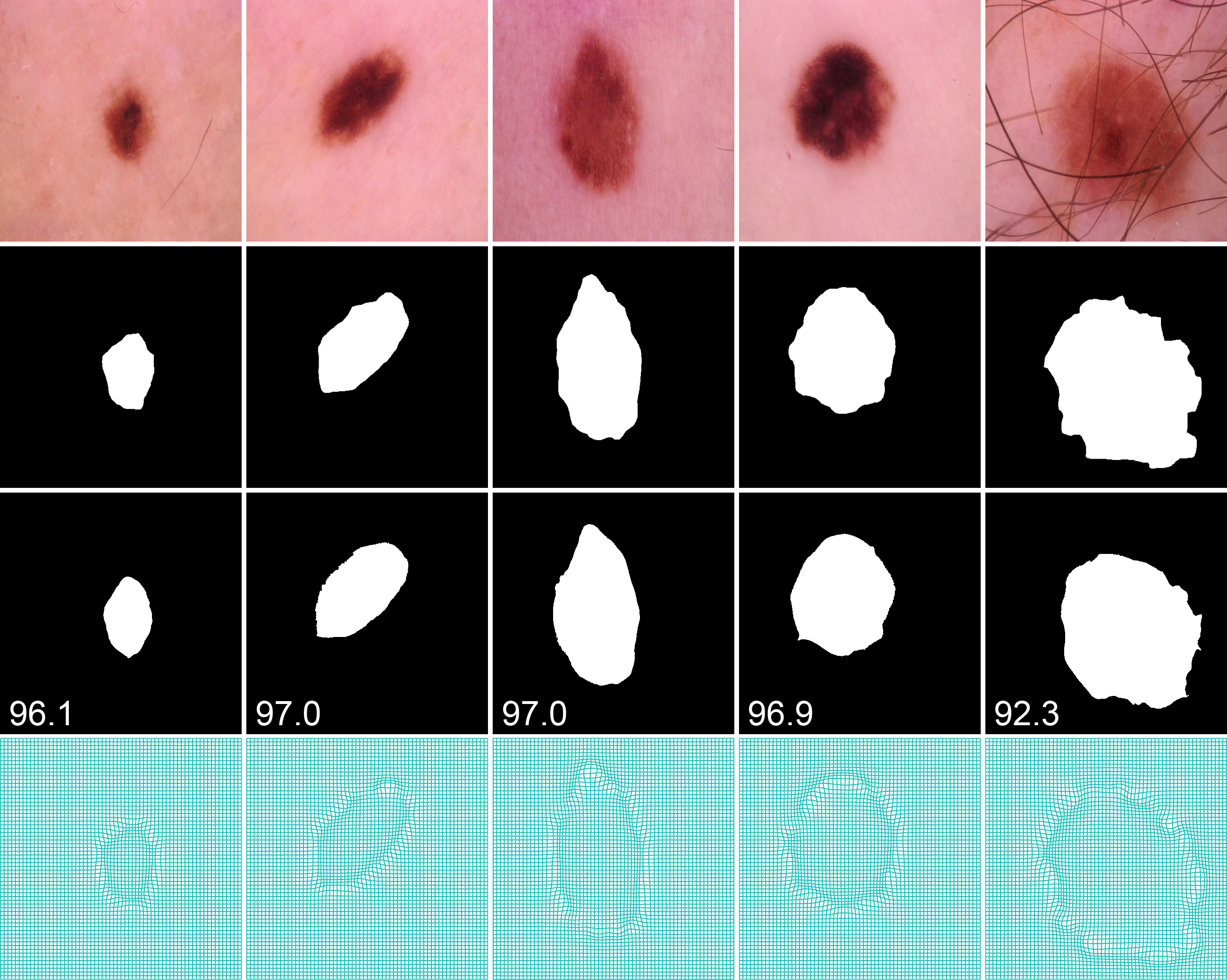}
    \caption{\han{The segmentation results for the selected HAM10000 dataset using the unsupervised TPSN (unTPSN). First row: the input image; second row: the ground truth; third row: the predicted mask by unTPSN; fourth row: the deformation maps.}}
    \label{fig:LesionBoundary}
\end{figure}

We also evaluated the performance of our proposed unsupervised image segmentation model, which does not require labeled masks for training. We conducted experiments on a selected Ham10000 dataset and predicted the segmentation mask using the unsupervised loss (Equation (\ref{eq:CV})). Taking a disk whose radius is a quarter of the image sizes as the template mask, the segmentation results are reported in TABLE \ref{tb:unTPSM}. Notably, our proposed unsupervised segmentation model takes only $0.02$ seconds to obtain the segmentation result upon successful training, while the optimization approach \cite{zhang2021topoconv} requires $2.87$ seconds on average to segment a single image. The qualitative results of unTPSN are presented in Figure \ref{fig:LesionBoundary}, where the second row shows the ground truth, the third row shows the segmentation results obtained by our proposed model {\color{red} and the last row shows their corresponding deformation maps.} The Dice scores are reported at the bottom left, indicating that the segmentation results achieved by our proposed method are satisfactory and comparable to those obtained by the optimization approach.

\begin{table}
\small
\centering
\begin{tabular}{ccccccc}
\toprule
Method  & Time (s)  & Dice      & HD        & Betti     & $\mathcal{L}_{Jac}$   & $\|\mu\|$\\\hline
Zhang's & 2.87      & 87.99     & \B{9.69}  & \B{1.00} & 0.00                   & 0.52\\
unTPSN  & \B{0.02}  & \B{88.15} & 9.94      & \B{1.00} & 0.00                   & 0.59\\\hline
\bottomrule
\end{tabular}
\captionof{table}{Quantitative comparison for unsupervised TPSN using the selected HAM10000 dataset.}
\label{tb:unTPSM}
\end{table}

\subsection{Multi-object Segmentation}

\newcolumntype{E}{>{\centering\arraybackslash}p{0.06\textwidth}}
\begin{table*}
\small
\centering
\begin{tabular}{E||p{0.13\textwidth}|E|E|E|E|E|E|E|E|E}
\toprule
Metric                  & Method        & Ave       & Spl       & Kid(R)    &Kid(L)     & Gall      & Liv       & Sto       & AG(R)     & AG(L)     \\\hline\hline
\multirow{7}{*}{Dice}   & UNet          & 62.90     & 82.59     & 86.89     & 82.39     & 36.06     & 87.96     & 52.86     & 42.87     & 31.64     \\
                        & wUNet         & 67.69     & 83.33     & 87.73     & 84.39     & 40.94     & 89.41     & 61.06     & 49.12     & 45.57     \\
                        & UNet+prior    & 82.07     & 87.60     & 88.79     & 87.26     & 78.87     & 90.08     & 87.62     & 65.89     & 70.43      \\
                        & wUNet+prior   & 82.93     & 89.05     & \B{89.45} & 87.76     & 81.61     & \B{91.27} & \B{87.90} & 66.42     & 70.00      \\
                        & TeTrIs    & 81.79     & 86.67     & 87.85     & 86.71     & \B{82.81} & 87.71     & 86.18     & 63.17     & 67.88      \\\cline{2-11}
                        & TPSN          & 82.28     & 88.45     & 88.96     & 88.95     & 81.07     & 89.99     & 86.48     & 64.98     & 69.34     \\
                        & mlTPSN        & \B{83.49} & \B{89.33} & 89.10     & \B{89.13} & 82.11     & 90.43     & 87.07     & \B{68.38} & \B{72.40} \\\hline\hline
\multirow{7}{*}{HD}     & UNet          & 10.15     & 8.14      & 13.50     & 6.62      & 6.27      & 14.71     & 18.78     & 6.58      & 6.65     \\
                        & wUNet         & 9.09      & 8.81      & 12.17     & 5.42      & 5.11      & 12.63     & 16.51     & 5.43      & 6.65      \\
                        & UNet+prior    & 5.52      & 5.46      & 3.87      & 3.88      & 3.79      & 11.50     & 6.24      & 3.69      & 5.78      \\
                        & TeTrIs    & 5.37      & 6.07      & 3.88      & 3.67      & 3.90      & 13.57     & 7.27      & 3.72      & 3.42      \\\cline{2-11}
                        & TPSN          & 5.23      & 5.28      & 3.54      & 3.54      & 3.31      & 10.43     & 7.06      & 3.47      & 3.23      \\
                        & mlTPSN        & \B{4.69}  & \B{5.13}  & \B{3.33}  & \B{3.47}  & \B{3.27}  & 9.97      & \B{6.12}  & \B{3.13}  & \B{3.06}  \\\hline\hline

\multirow{7}{*}{Betti}  & UNet          & 3.27      & 1.60      & 5.20      & 2.00      & 1.60      & 5.60      & 6.20      & 1.60      & 2.40      \\
                        & DoubleUNet    & 3.00      & 1.60      & 4.60      & 2.00      & 1.20      & 4.80      & 7.00      & 1.20      & 1.60      \\
                        & UNet+prior    & 2.08      & 1.20      & \B{1.00}  & \B{1.00}  & 1.40      & 7.40      & 2.00      & 1.40      & 1.20      \\
                        & wUNet+prior   & 1.63      & 1.20      & \B{1.00}  & \B{1.00}  & 1.40      & 4.40      & 1.60      & 1.20      & 1.20      \\
                        & TeTrIs    & 1.13      & 1.20      & \B{1.00}  & 1.63      & 1.20      & 1.20      & \B{1.00}  & 1.20      & 1.40      \\\cline{2-11}
                        & TPSN          & \B{1.00}  & \B{1.00}  & \B{1.00}  & \B{1.00}  & \B{1.00}  & \B{1.00}  & \B{1.00}  & \B{1.00}  & \B{1.00}     \\
                        & mlTPSN        & \B{1.00}  & \B{1.00}  & \B{1.00}  & \B{1.00}  & \B{1.00}  & \B{1.00}  & \B{1.00}  & \B{1.00}  & \B{1.00}     \\\hline\hline
                            
\multirow{3}{*}{$\mathcal{L}_{Jac}$}
                        & TeTrIs        & 0.84      & 0.72      & 0.57      & 0.38      & 0.67      & 3.07      & 0.86      & 0.37      & 0.22     \\\cline{2-11}
                        & TPSN          & 0.00      & 0.00      & 0.00      & 0.00      & 0.00      & 0.00      & 0.00      & 0.00      & 0.00     \\
                        & mlTPSN        & 0.00      & 0.00      & 0.00      & 0.00      & 0.00      & 0.00      & 0.00      & 0.00      & 0.00     \\\hline
\bottomrule
\end{tabular}
\captionof{table}{Quantitative results of multi-object segmentation for the BCTV dataset. Ave: averaged over all organs, Spl: spleen, Kid(R): right kidney, Kid(L): left kidney, Gall: gallbladder, Liv: liver, Sto: stomach, AG(R): right adrenal gland, AG(L): left adrenal gland}
\label{tb:multiclass}
\end{table*}
All of the previous experiments focus on the segmentation of a single structure in the image. In this subsection, we examine the performance of our proposed model for multi-object segmentation, which is a fundamental task in medical imaging. We used the BTCV dataset from the MICCAI 2015 Multi-Atlas Abdomen Labeling Challenge\cite{btcv2015} for this experiment. The template masks for multiple objects used in this experiment were the bounding boxes for the organs, which can be extracted by detection networks \cite{baumgartner2021nndetection}. To provide the same prior information for U-Net and double U-Net, we concatenated the template masks with the original image as the input. 

The quantitative results, as shown in Table \ref{tb:multiclass}, highlight the effectiveness of our method, which surpasses all other models. When using pixel-wise classification approaches, methods like UNet and doubleUNet face challenges in accurately preserving topology for all organs, even with prior input information. Notably, TeTrIs \cite{lee2019tetris} fails to maintain topology, as evidenced by the Betti number errors. Additionally, the mapping generated by TeTrIs is not guaranteed to be bijective, as indicated by the non-zero ReLU-Jacobian errors. This emphasizes the inadequacy of relying solely on Laplacian regularization to ensure mapping bijectivity. In contrast, our approach consistently produces segmentation results with correct topology and achieves superior Hausdorff distances for most organs.

\begin{figure}
    \centering
    \includegraphics[width = 0.68\textwidth]{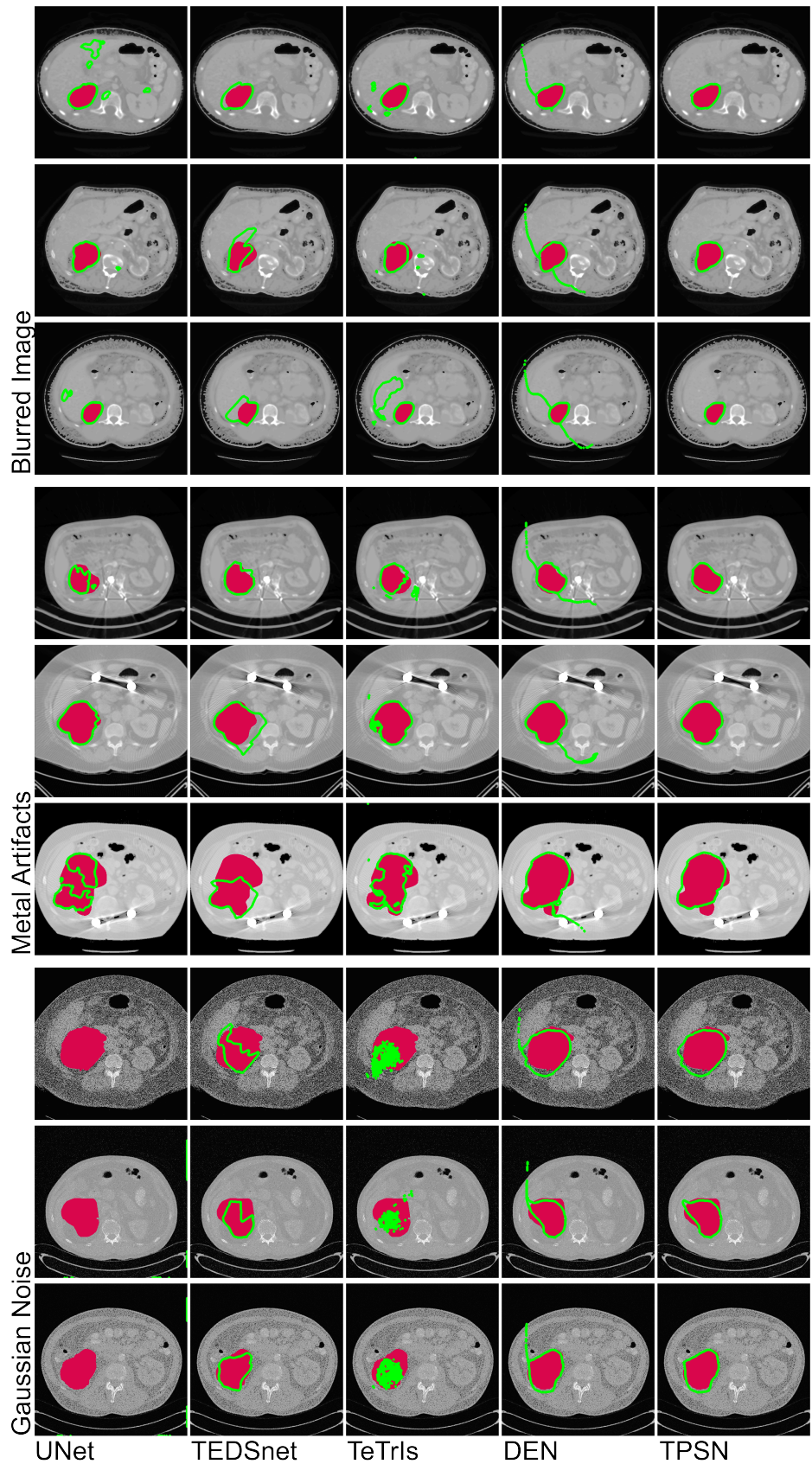}
    \caption{Visual comparison of Unet, TEDSnet, TeTrIs, and our proposed DEN (TPSN without BAM) and TPSN. The images are degraded by: image blurs, metal artifacts, and noises. The predicted masks obtained methods are indicated by green contour while the ground truth is in red.}
    \label{fig:kits_noise}
\end{figure}

\subsection{Segmentation for Corrupted Data}

Our framework incorporates two crucial components: the Deformation Estimation Network (DEN) and the Beltrami Adjustment Module (BAM). DEN is trained on data to predict the mapping required to warp the template mask, while BAM serves as a post-processing module to ensure that the mapping generated by DEN is fully bijective. Under normal circumstances, DEN is effective in producing a bijective mapping that can accurately warp the template mask into the target region while preserving topology. However, it may encounter challenges in maintaining this essential property when presented with input data that falls outside the distribution of the training dataset. For example, if a network is trained exclusively on high-quality images, it may struggle to generate a bijective mapping for a low-quality image with significant noise. In such scenarios, BAM comes into play by adjusting the mapping $f$ produced by DEN to a bijective mapping $\Tilde{f}$ while minimizing the discrepancy between the newly warped mask $M_{temp} \circ \Tilde{f}$ and the original warped mask $M_{temp} \circ {f}$ generated by DEN. As a result, our framework consistently produces segmentation results that preserve topology, even under challenging circumstances.

To evaluate our method, we synthesized corrupted data that contains three types of image degradations commonly encountered in medical imaging. The first type is blurred images, which may result from unstable equipment settings or patient movement during imaging. The second type is images with Gaussian noise, which may be generated by signal noise in imaging machines. The third type is images with metal artifacts that may arise from in-vivo medical devices used by patients. 

In Figure \ref{fig:kits_noise}, we present the segmentation results obtained by our proposed framework and other state-of-the-art deformation-based segmentation models as well as the backbone UNet. The first two rows show the segmentation results of four blurry images. The red regions represent the ground truth, while the green contour shows the predicted mask obtained by different models. Our proposed method with the BAM yields the best results without topological errors. The segmentation results obtained by only DEN module without BAM may still contain topological errors. The third and fourth rows depict the segmentation results of four images with metal artifacts. Once again, our proposed method with BAM performs the best, while other methods produce mis-segmentation results and contain topological errors. The last two rows show the segmentation results of four noisy images. Our proposed method with BAM yields the best results without topological errors. These results demonstrate the capability of our proposed model to handle corrupted images.

\section{Conclusion}
In this work, we propose a novel learning-based segmentation framework for topology-preserving image segmentation, namely the Topology-Preserving Segmentation Network (TPSN). The image to be segmented is first inputted into the Deformation Estimation Network. As a deformable model, the Deformation Estimation Network completes the segmentation task by generating a mapping using an encoder-decoder architecture that transforms the template mask into the region of interest. The mapping is regularized by the proposed ReLU Jacobian regularization, which encourages the orientation-preserving property of the mapping. The mapping is further fed into the Beltrami Adjustment Module to ensure bijectivity, thereby preserving the topology of the transformed mask. Our proposed learning model can be either a supervised model or extended to an unsupervised segmentation model using an intensity-based unsupervised loss function.

Extensive experiments were conducted to demonstrate the effectiveness of our proposed framework. We conducted self-ablation studies to analyze the effects of various parameters in the framework. To evaluate the capability of our model in different dimensions, we performed experiments on both 2D planar images and 3D volumetric images. We also conducted experiments to segment structures with different topologies. Compared with other state-of-the-art methods, our proposed method showed comparable results by Dice score and often outperformed others by geometric measures like Hausdorff distance. Furthermore, with the carefully designed ReLU Jacobian term and Beltrami Adjustment Module, our method consistently predicted segmentation masks with the correct topology.

In future work, we plan to investigate the use of more geometric constraints in our topology-preserving segmentation network. Specifically, we plan to explore the use of convexity\cite{zhang2021topoconv} as a constraint in the segmentation process. Additionally, we will consider incorporating some explicit geometry features such as the shape signature\cite{lin2022harmonic} to further improve the accuracy of the segmentation. Moreover, while our current TPSN method provides accurate segmentation results, it is limited by the fixed topology of the input template mask. In future work, we plan to extend our method to incorporate interactive segmentation\cite{spencer2018parameter}, allowing users to modify the topology of the segmentation mask interactively in real-time. This will enable more flexible and user-friendly segmentation.

\section*{Acknowledgment}
This work was supported by HKRGC GRF (Project ID: 14307622), and Hong Kong Centre for Cerebro- Cardiovascular Health Engineering (COCHE).

\bibliographystyle{plain}  
\bibliography{reference}

\end{document}